\documentclass[journal,transmag]{IEEEtran}
\usepackage{amsmath,amsfonts}
\usepackage{array}
\usepackage{textcomp}
\usepackage{stfloats}
\usepackage{url}
\usepackage{verbatim}
\usepackage{graphicx}
\usepackage{subfigure}
\usepackage{caption}
\usepackage{tabularx}
\usepackage{multirow}
\usepackage{makecell}
\usepackage{booktabs}
\usepackage{array} 
\usepackage{setspace}
\usepackage{ulem}
\usepackage{amsmath}

\usepackage[numbers,sort&compress]{natbib} 

\usepackage[linesnumbered, ruled]{algorithm2e}
\renewcommand{\emph}[1]{\textit{#1}}
\usepackage{hyperref}

\usepackage{color}

\usepackage{graphicx} 
\usepackage{lipsum} 

\usepackage{subcaption}
\usepackage{lipsum} 

\usepackage[table]{xcolor} 

\ifCLASSINFOpdf
\else
\fi
\begin{document}
%
\title{CSPENet: Contour-Aware and Saliency Priors Embedding Network for Infrared Small Target Detection}


\author{\IEEEauthorblockN{Jiakun Deng\IEEEauthorrefmark{†},
Kexuan Li\IEEEauthorrefmark{†},
Xingye Cui\IEEEauthorrefmark{}, 
Jiaxuan Li\IEEEauthorrefmark{}, 
Chang Long\IEEEauthorrefmark{}, 
Tian Pu\IEEEauthorrefmark{*}, and \\
Zhenming Peng\IEEEauthorrefmark{*},~\IEEEmembership{Member,~IEEE}}
\thanks{\IEEEauthorrefmark{†}These authors contributed equally to this work and should be considered co-first authors.}
\thanks{This work was supported by Natural Science Foundation of Sichuan Province of China (Grant No.2025ZNSFSC0522) and partially supported by National Natural Science Foundation of China (Grant No.61571096). (Corresponding authors: Zhenming Peng; Tian Pu.)}
\thanks{Jiakun Deng, Kexuan Li, Xingye Cui, Jiaxuan Li, Chang Long, Tian Pu and Zhenming Peng are with the School of Information and Communication Engineering and the Laboratory of Imaging Detection and Intelligent Perception, University of Electronic Science and Technology of China, Chengdu 610054, China
(email: dengjiakun@std.uestc.edu.cn; kexuanli@std.uestc.edu.cn; cxy011211@163.com; 202422011409@std.uestc.edu.cn; lc243265379@gmail.com; putian@uestc.edu.cn; zmpeng@uestc.edu.cn)}

}

\markboth{Journal of \LaTeX\ Class Files,~Vol.~14, No.~8, April~2025}%
{Shell \MakeLowercase{\textit{et al.}}: Bare Demo of IEEEtran.cls for IEEE Transactions on Magnetics Journals}
\maketitle

\IEEEdisplaynontitleabstractindextext

%
\IEEEpeerreviewmaketitle

\begin{abstract}
\textbf{\textit{Abstract—}}Infrared small target detection (ISTD) plays a critical role in a wide range of civilian and military applications. Existing methods suffer from deficiencies in the localization of dim targets and the perception of contour information under dense clutter environments, severely limiting their detection performance. To tackle these issues, we propose a contour-aware and saliency priors embedding network (CSPENet) for ISTD. We first design a surround-convergent prior extraction module (SCPEM) that effectively captures the intrinsic characteristic of target contour pixel gradients converging toward their center.  This module concurrently extracts two collaborative  priors: a boosted saliency prior for accurate target localization and multi-scale structural priors for comprehensively enriching contour detail representation. Building upon this, we propose a dual-branch priors embedding architecture (DBPEA) that establishes differentiated feature fusion pathways, embedding these two priors  at optimal network positions to achieve  performance enhancement. Finally, we develop an attention-guided feature enhancement module (AGFEM)  to refine feature representations and improve saliency estimation accuracy. Experimental results on public datasets NUDT-SIRST, IRSTD-1k, and NUAA-SIRST demonstrate that our CSPENet outperforms other state-of-the-art methods in detection performance. The code is available at https://github.com/IDIP2025/CSPENet.

\end{abstract}


\section{Introduction}
%
%
%
%
\IEEEPARstart{I}{nfrared} small target detection (ISTD), as one of the core technologies of all-weather optoelectronic sensing systems, has irreplaceable military and civilian value in precision guidance, strategic early warning, remote sensing monitoring, and related fields \cite{abruzzi2025internal}. Due to the inherent limitations of the imaging process, infrared images have three problems: 1) Ultra-low pixel occupancy: Infrared targets typically cover less than 0.15$\%$ of the image size, resulting in the sparsity of the target. 2) Degraded signal-to-noise ratio (SNR): Dim targets and overwhelming background clutter contribute to a poor SNR. 3) Shape variation: Unstructured geometries resulting from variations in imaging distance, target motion, and environmental factors lead to significant changes in target appearance \cite{zhang2025istd}. These problems have rendered ISTD a persistent research challenge, prompting extensive research efforts over the past few decades.

 Early research efforts in ISTD predominantly relied on model-driven approaches that leveraged physical priors for target detection. Representative methods include: 1) Filter-based methods  \cite{zhang2025mirsam}, \cite{deshpande1999max},  which assume  that the background has spatial continuity and the target is abrupt singular points. 2) Human visual system (HVS)-inspired saliency detection-based methods \cite{chen2013local}, \cite{wei2016multiscale}, which leverage contrast differences between targets and local backgrounds to detect salient targets. 3) Low-rank and sparse decomposition frameworks-based methods \cite{zhang2019infrared}, \cite{yi2023spatial}, which formulate detection as a matrix decomposition problem by enforcing background low-rankness and target sparsity. While these methods offer strong interpretability and eliminate the need for large-scale training data, their performance is critically dependent on the validity of the prior assumptions. Consequently, they suffer from inherent limitations such as high parameter sensitivity and inadequate generalization in complex scenarios \cite{zhang2025istd}.

\begin{figure}
 \centering
 \includegraphics[scale=0.45]{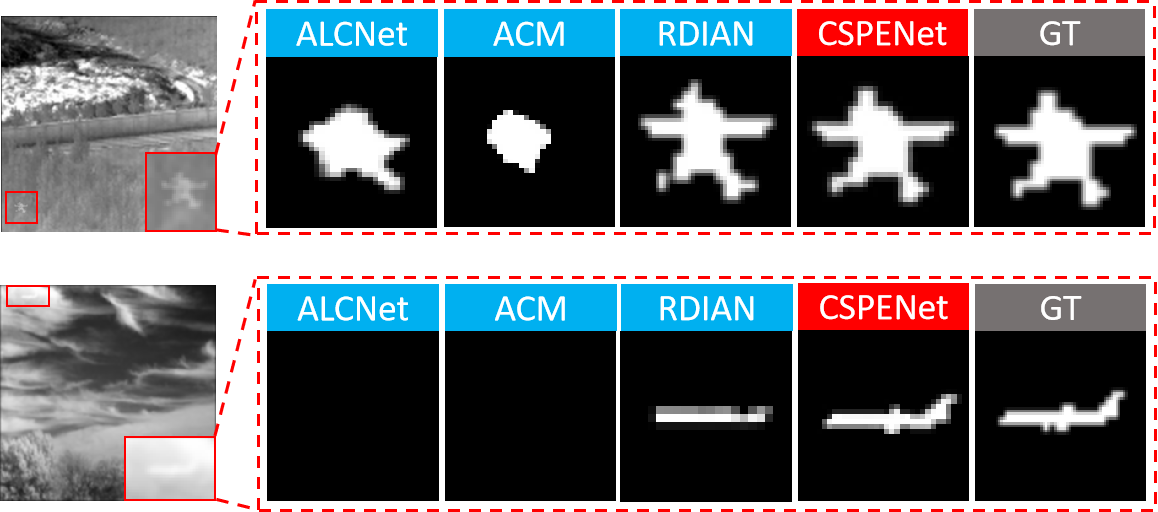}
 \caption{\small{ Visual comparison. Segmentation results of complex-structured targets (zoomed) in dense clutter environment obtained by different methods.\label{VC}}}
 \label{VC}
\end{figure}

In recent years, the advent of a large number of annotated datasets has facilitated the development of deep learning (DL)-based ISTD methods \cite{wu2024infrared}, \cite{yuan2024sctransnet}, \cite{zang2025dcanet}. These methods have outperformed traditional model-driven approaches in terms of detection accuracy and false alarm rate control, thanks to their end-to-end network architectures and backpropagation optimization mechanisms. However, current DL-based methods still encounter two significant technical challenges: 1) The inherent reliance on large-scale training data restricts the model's ability to focus on localized patterns, leading to inaccurate target localization in dense clutter environments \cite{xu2024hcf}. 2) Deep features are subject to semantic degradation since repeated convolution and pooling operations cause deep neural networks to lose low-level spatial details, thus failing to effectively preserve small target contours \cite{tishby2000information}. As shown in Fig. \ref{VC}, DL-based models without embedded structural priors demonstrate inaccurate segmentation performance and frequently fail to detect targets in dense clutter scenarios.

\begin{figure}[htbp]
 \centering
 \includegraphics[scale=0.25]{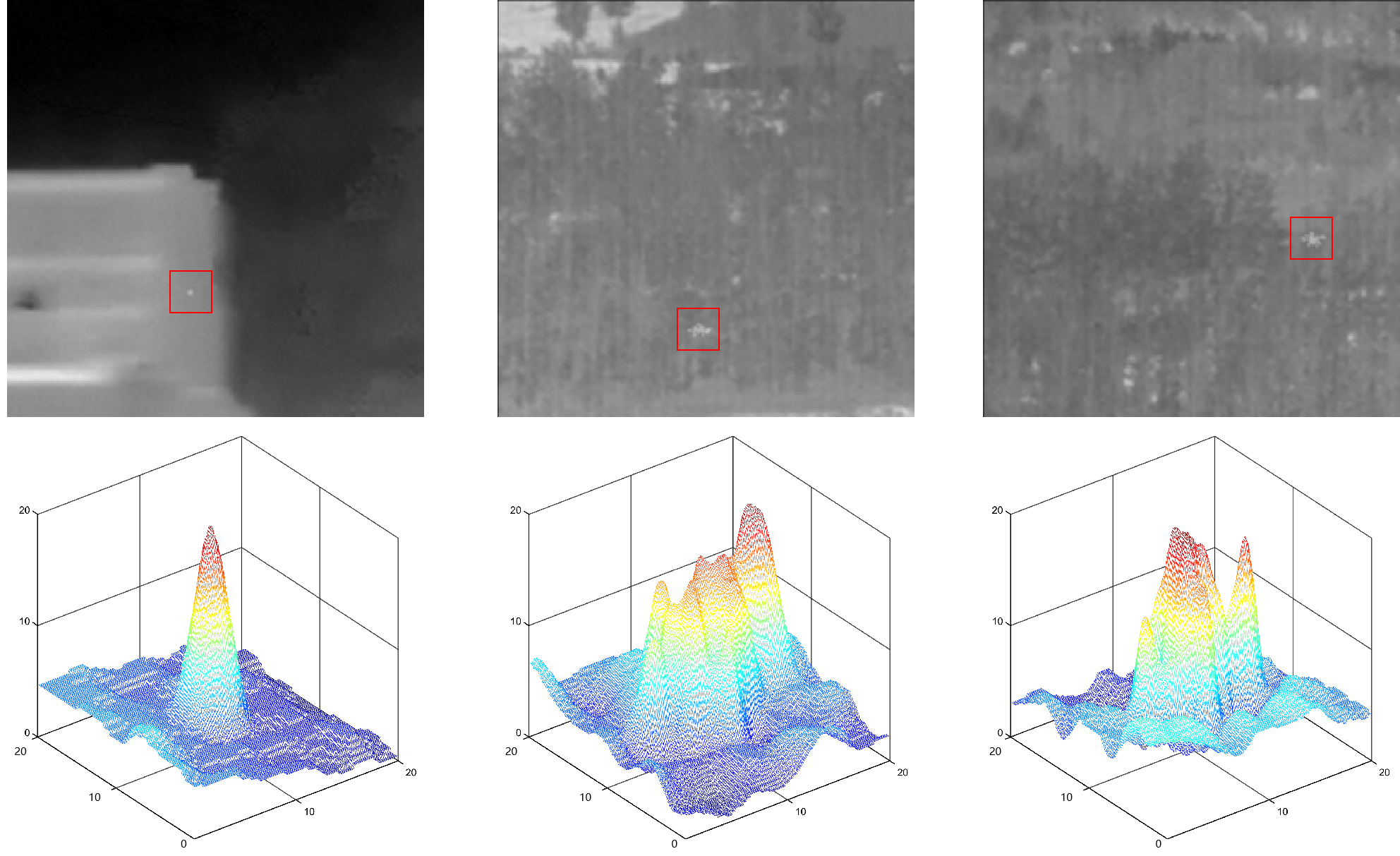}
 \caption{\small{Infrared small target image (upper) and its corresponding 3D local structural characterization visualization (lower).}}
 \label{introduction}
\end{figure}

Several existing studies have endeavored to integrate prior knowledge, such as HVS characteristics, into DL frameworks to enhance target localization and detail description capabilities \cite{dai2021attentional}, \cite{yu2022infrared}, \cite{wang2025paying}. However, the lack of consideration for the structural priors of infrared targets and the difficulty in fusing physical properties with deep representations using traditional cascaded strategies limit the improvement of model performance. Through the analysis of extensive infrared target imaging data, we observe that target contours exhibit a significant surround-convergent characteristic, as illustrated in Fig. \ref{introduction}. Specifically, the radiation intensity in the central region is higher than that in the surrounding edge regions, and the gradient directions of most edge pixels converge toward the target center. Leveraging this structural characteristic, we can coarsely localize dim targets, thereby effectively suppressing false alarms in complex scenes. Furthermore, the surround-contour structural priors enable finer-grained representation of the target's edge details. 

In this paper, we propose a contour-aware and saliency priors embedding network (CSPENet) for ISTD, which effectively addresses the limitations of existing methods in target localization and contour detail representation by explicitly perceiving surround-convergence priors from target contours.  Specifically, we design a surround-convergent prior extraction module (SCPEM), which employs multidirectional gradient magnitude calculation blocks (MGMCBs) to extract two collaborative convergent prior (CP) components: 1) CP1, a boosted saliency structure prior for precise target localization, and 2) CP2, multi-scale surround-convergent priors encoding to enhance contour hierarchical details. These components are differentially  embedded into the deep network via a dual-branch priors embedding architecture (DBPEA): CP1 enhances initial target perception by jointly encoding with the input image, while CP2 dynamically interact with deep semantic features through a cross-hierarchical knowledge-infused module (CHKIM),  thereby mitigating the degradation of contour details. Finally, an attention-guided feature enhancement module (AGFEM) adaptively fuses multi-layer features, enabling robust detection of weak and irregularly shaped targets. Subsequent experiments have demonstrated that our CSPENet achieves superior detection results on multiple datasets compared to other state-of-the-art (SOTA) algorithms.

In summary, our work makes four main contributions.

1) We propose CSPENet, a novel ISTD framework that embeds contour-aware and saliency priors for infrared targets into an end-to-end detection frameworks. This approach enhances the model's performance in precise target localization and detailed contour preservation under  dense and cluttered environments.

2) We develop a structure prior extractor SCPEM to generate two collaborative priors -- a boosted saliency map CP1 for precise spatial localization and  multi-scale CP2 for contour detail preservation -- effectively capturing both target saliency and structural characteristics.

3) We propose DBPEA to effectively embed CP components through dual pathways: CP1 directly encodes with input imagery to boost initial localization and spatial awareness, while CHKIM adaptively merges CP2 with deep features to restore semantic details and maintain structural integrity.

4) We construct AGFEM to dynamically optimize multi-layer features through attention mechanisms, significantly boosting detection performance for challenging weak or irregular targets.

\section{RELATED WORK}
\subsection{Infrared Small Target Detection}
\textit{1) Model-Driven Methods:} 
Existing model-driven ISTD methods can be categorized into three groups: filter-based methods, HVS-based methods, and low-rank sparse decomposition-based methods. i) Filter-based methods (e.g., Top-hat  \cite{bai2010infrared}, Max-Median \cite{deshpande1999max}, and 2-D adaptive TDLMS filters \cite{soni1993performance}) alongside Gaussian-derived techniques (Laplacian of Gaussian \cite{kim2009small}, Difference of Gaussian \cite{wang2012infrared}) operate under the background consistency assumption, where targets disrupt local gray-level uniformity \cite{xu2023infrared}. Recent advancements have mitigated limitations such as background clutter \cite{shi2017high}, \cite{zhang2023infrared}. ii) HVS-based methods leverage local contrast mechanisms, evolving from the foundational LCM \cite{chen2013local}  to improved variants like RLCM \cite{han2018infrared}, TLLCM \cite{han2019local}, and ADMD \cite{moradi2020fast}. Recent advancements include DL-enhanced frameworks such as LELCM \cite{yang2024label}. iii) Low-rank sparse decomposition methods, initiated by IPI \cite{gao2013infrared}, separate sparse targets from low-rank backgrounds through tensor modeling approaches like PSTNN \cite{zhang2019infrared} and tensor train/ring decomposition \cite{wu2023infrared}, as well as optimization improvements incorporating spatial-temporal regularization \cite{yi2023spatial}. However, due to their reliance on handcrafted shallow features, these methods often suffer from high parameter sensitivity and limited generalization capabilities in complex scenarios.

\textit{2) Deep Learning Methods:} 
In recent years, DL-based methods, leveraging their adaptive feature extraction, have gradually become the dominant research paradigm in this field.

Early works were mostly based on U-Net and its variants, such as ACM \cite{dai2021asymmetric}, DNANet \cite{li2022dense}, UIUNet \cite{wu2022uiu}, and AMFU-net \cite{chung2023lightweight}. Recently, Zhang et al. \cite{zhang2024mdigcnet} proposed MDIGCNet, which introduced the integrated differential convolution (IDConv) module based on the U-Net architecture to extract richer image features. Zang et al. \cite{zang2025dcanet} designed a dense pixel contrastive learning module aiming at capturing more features of targets with similar sizes.
To break through the limitations of local receptive fields, researchers have explored the collaborative modeling mechanism of Vision Transformers (ViT) and convolutional neural networks (CNNs). Yuan et al. \cite{yuan2024sctransnet} developed SCTransNet to achieve cross-scale semantic association through a spatial channel cross-attention module. Wu et al. \cite{wu2023mtu} designed MTU-Net to capture complementary global context and local detail features through a hybrid encoder of ViT and CNN. Zhu et al. \cite{zhu2024towards} proposed GSTUnet that employs an edge-guided attention mechanism to selectively enhance the response in target regions. Wu et al. \cite{wu2024infrared} proposed STASPPNet to optimize multi-scale feature representation using Swin Transformer with serial dilated convolutions.

Despite the significant improvement in detection performance demonstrated by DL methods, pure DL models are prone to overfitting and cross-scene generalization bottlenecks due to the high cost of infrared data annotation and inherent noise interference \cite{xu2024hcf}. To address this problem, researchers have attempted to jointly design classical model-driven frameworks with DL networks. For example, the mathematical modeling of traditional filters has been transformed into learnable convolutional kernel constraints, with frequency-domain priors guiding feature extraction directions \cite{dai2021attentional}, \cite{yu2022infrared}. Additionally, the optimization objectives of low-rank sparse decomposition have been embedded into the loss function to construct regularization constraint modules based on matrix recovery theory, suppressing redundant background components \cite{wu2024rpcanet}, \cite{kou2023infrared}. However, existing methods mostly adopt loose, auxiliary combination strategies, such as simply concatenating the preprocessed results from traditional algorithms with CNN features or stacking shallow physical constraints in the loss function, failing to achieve deep integration between model-driven mechanisms and DL architectures. 


\begin{figure*}
 \centering
 \includegraphics[width=1\linewidth]{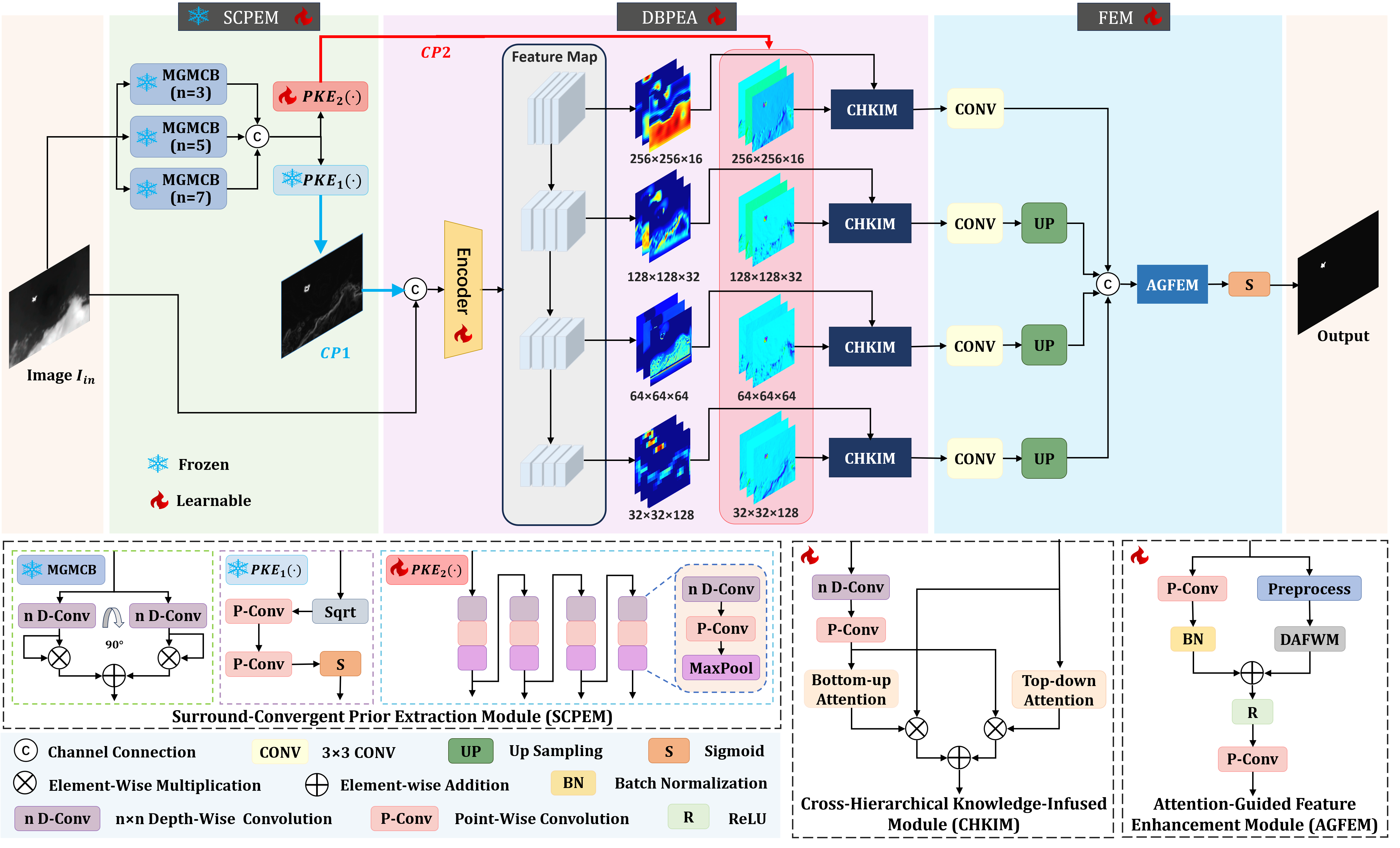}
 \caption{ \small{The overall architecture of the proposed CSPENet for infrared small target detection. 1) SCPEM: The input image is first fed into SCPEM to obtain two types of CP components. 2) DBPEA: The two types of CP components are embedded into the deep learning network through DBPEA, where CP1 is jointly encoded with the input image via channel concatenation, and CP2 dynamically interacts with the corresponding deep semantic features through CHKIM. 3) FEM: The fused features are convolved and upsampled to a uniform scale and concatenated in the channel dimension, and then AGFEM is used to adaptively enhance the features from different layers to produce the final predicted image.}}
 \label{backbone}
 \vspace{-0.5cm}
\end{figure*}

\subsection{Prior-Embedded Deep Detectors}

In the field of target detection, the collaborative reasoning mechanism embedding  prior knowledge into DL-based models has become a research focus in recent years. Compared with the conventional DL detection paradigm, the embedding of structured prior information introduces interpretable cognitive mechanisms for deep neural networks, effectively enhancing model robustness against occlusion, small targets, and complex scenes \cite{xu2019reasoning}. Existing research mainly explores three dimensions: multimodal knowledge representation, spatial relationship modeling, and scene constraint optimization. In terms of knowledge representation, Fang et al. \cite{fang2017object} constructed an interaction mechanism between knowledge graphs and visual features to significantly improve target recall rates through a hierarchical knowledge infusion strategy. Jiang et al. \cite{jiang2018hybrid} designed a routing module that combines multiple types of prior knowledge to introduce semantic relationships into visual features. Xu et al. \cite{xu2019spatial} innovatively proposed the spatially-aware graph convolutional network (SGRN) to jointly encode target semantics and spatial relationships. Yang et al. \cite{yang2019prior} established a knowledge-aware target detection framework to effectively solve contextual modeling challenges in large-scale detection tasks. Cheng et al. \cite{cheng2021target} built a context-aware detection framework from the perspective of scene semantic constraints to use the relationships between scenes and objects as specific prior knowledge to improve detection accuracy. Mo et al. \cite{mo2024novel} integrated geometric constraints and object co-occurrence prior knowledge to design a graph reasoning module that significantly enhances the model's deductive reasoning ability for complex target relationships.

In the ISTD field, the inherent feature sparsity of small targets \cite{ying2023mapping}, \cite{wang2019carafe}  and the limited scale of training samples make conventional DL-based models fall into the dilemma of insufficient representation learning, weakening their capability of distinguishing targets and complex background interferences. To address this issue, researchers have attempted to embed prior knowledge into DL frameworks. For example, ALCNet \cite{dai2021attentional} and the MLCL \cite{yu2022infrared} module embed the multi-scale patch-based contrast measure (MPCM) algorithm into the skip connections of DL networks through dilated convolution modules. Wang et al. \cite{wang2025paying} proposed a local contrast perception mechanism that jointly optimizes target enhancement and background suppression using region saliency priors. Zhao et al. \cite{zhao2024multi} infused the high-frequency directional features of infrared small targets as domain-specific prior knowledge into neural networks to alleviate the limitations of small target appearance representation. ISNet  \cite{zhang2022isnet} designed an edge-aware module based on Taylor finite difference to enhance shape prior modeling capabilities. DMEF-net \cite{ma2023dmef} integrated radiance distribution and motion characteristic priors to design a multi-physical feature compensation mechanism to mitigate the data sparsity contradiction. However, existing methods struggle to effectively extract structural priors from infrared targets and fail to adequately integrate their physical attributes with deep learning representations. This leads to an inherent trade-off between target localization accuracy and detailed structural description, ultimately limiting further improvements in detection performance. 



\section{METHODOLOGY}

\subsection{Overall Architecture}
As illustrated in Fig. \ref{backbone}, the CSPENet framework consists of three components: SCPEM, DBPEA, and feature enhancement module (FEM). SCPEM extracts structural prior components from the input image, specifically generating two types of CP components: one for localization enhancement (CP1) and the other for multi-scale contour detail supplementation (CP2). Subsequently, DBPEA differentially embeds CP1 and CP2 through a dual-path architecture to enhance feature representation, effectively balancing localization accuracy and contour detail preservation. Finally, in FEM, features from all layers are upsampled to the same resolution and refined via AGFEM, producing a more precise target saliency map.

\subsection{SCPEM}
\label{The Structure Prior Extraction Module}

\textit{1) Motivation:} 
The inherent thermal diffusion characteristics of infrared imaging induce distinct luminance variations between target centers and peripheral regions. Embedding such local structural priors enables ISTD systems to effectively perceive real target areas.

Existing studies, such as local binary patterns (LBP), have demonstrated effectiveness in representing invariant features for ISTD tasks \cite{wu2023repisd}, \cite{yu2020searching}, \cite{ying2022local}. However, these methods are limited to capturing contrast information along horizontal, vertical, and diagonal directions, making them inadequate for characterizing complex-structured targets.
Recognizing that multi-directional contrast information can significantly enhance detection performance for complex-structured targets, while solely relying on contrast priors lacks detailed contour representation, we propose SCPEM. This module utilizes multi-directional Gaussian derivative kernels to capture the target's surround-convergence characteristics beyond simple contrast. The multi-directional Gaussian derivative kernels provide two key benefits: first, they enhance localization performance for complex targets by leveraging contrast information from multiple directions; second, they capture contour details to improve target segmentation accuracy. The architecture consists of three-scale MGMCBs and two CP extraction modules. 

\begin{figure*}
 \centering
 \includegraphics[width=1\linewidth]{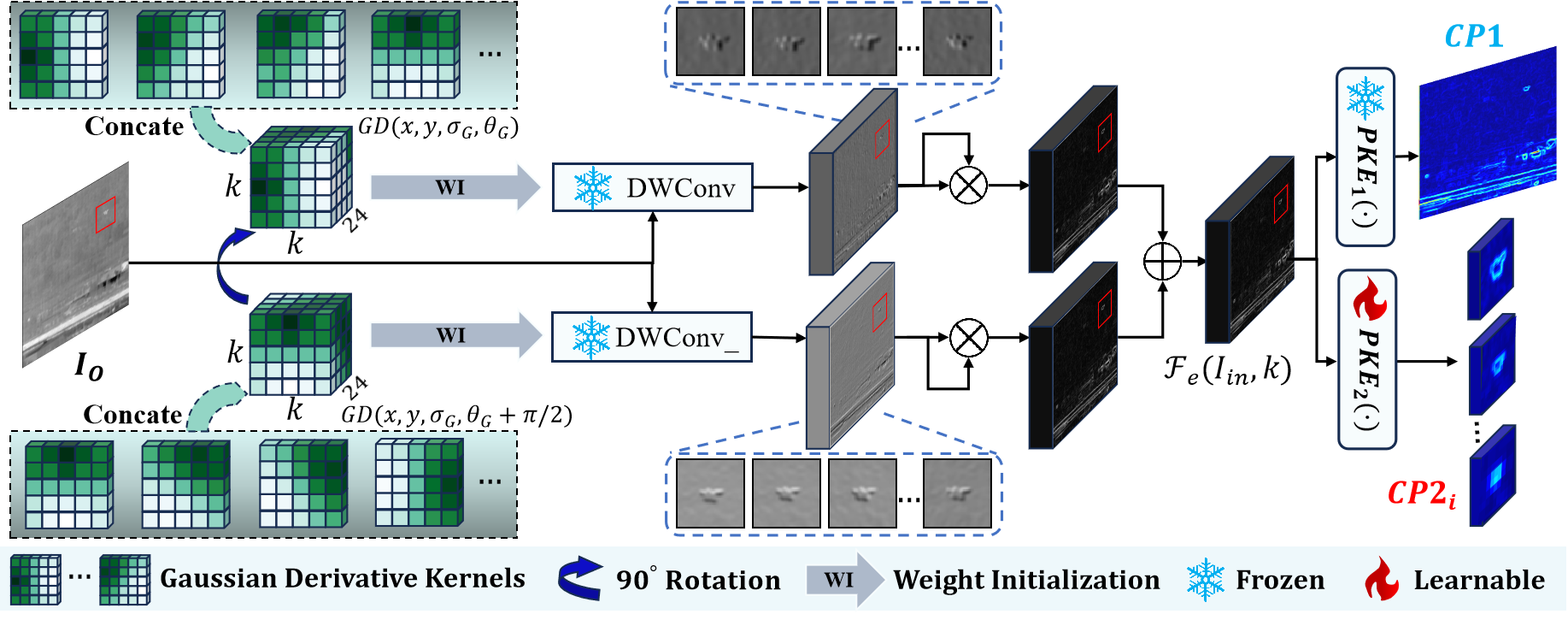}
 \caption{ \small{The architecture of MGMCB.}}
 \label{SCPEM}
 \vspace{-0.5cm}
\end{figure*}

\textit{2) MGMCB:}
The block is derived by unfolding the traditional gradient magnitude computation process into a deep network architecture. The core operation employs Gaussian derivative kernels $GD\left(x, y, \sigma_{G}, \theta_{G}\right)$ to extract oriented local structures, with the kernel defined as: 
\begin{equation}
	\centering
	\begin{array}{l}
		GD\left(x, y, \sigma_{G}, \theta_{G}\right)=-\dfrac{G\left(x, y, \sigma_{G}\right)}{\sigma_{G}^{2}}\left(x \cos \theta_{G}+y \sin \theta_{G}\right)
	\end{array}
	\label{eq3_2_1}
\end{equation}
where $G\left(x, y, \sigma_{G}\right)=\left(1 / \sqrt{2 \pi \sigma_{G}^{2}}\right) \mathrm{e}^{-\left(x^{2}+y^{2}\right) / 2 \sigma_{G}^{2}}$ denotes the  Gaussian  kernels with scale parameter $\sigma_{G}$, and $\theta_{G}$ denotes the partial derivative direction.

The calculation of the gradient magnitude $I_{M}\left(x, y, \sigma_{G}, \theta_{G}\right)$ combines the responses in orthogonal directions:
\begin{equation}
	\centering
	\begin{array}{l}
		I_{M}\left(x, y, \sigma_{G}, \theta_{G}\right)=\left( G D\left(x, y, \sigma_{G}, \theta_{G}\right) \otimes I_{in}(x, y)\right)^{2}\\ \quad \quad \quad \quad +
        \left( G D\left(x, y, \sigma_{G}, \theta_{G}+\pi/2\right) \otimes I_{in}(x, y)\right)^{2}
	\end{array}
	\label{eq3_2_2}
\end{equation}
where $ I_{in}\in \mathbb{R} ^{W\times H\times 1}$ is the input image, and $\otimes$ denotes convolution. $W$ and $H$ represent the width and height of the input respectively. To effectively capture the surround-convergence characteristics of diverse targets (particularly irregular ones), we construct the MGMCB output by stacking gradient magnitude feature maps from 24 discrete orientations [0°, 15°, 30°, ..., 345°].

For neural network implementation, we reformulate this process using depth-wise convolutions to obtain the gradient magnitude tensor $\mathcal{F} _{e}(I_{in} ,k)$:
\begin{equation}
	\centering
	\begin{aligned}
		\mathcal{F} _{e}(I_{in} ,k) =\mathrm{DWConv}(I_{in},k)^{2} +\mathrm{DWConv}\rule{0.5em}{0.4pt}(I_{in},k)^{2}
	\end{aligned}
	\label{eq3_2_7}
\end{equation}
where both $\mathrm{DWConv}( \cdot)$ and $\mathrm{DWConv}\rule{0.5em}{0.4pt}( \cdot)$  represent complementary depth-wise convolutional operations that preserve the original gradient computation properties while enabling efficient network integration. $k$ represents the size of the convolution kernel in the depth-wise convolution operation. Fig. \ref{SCPEM} presents the architecture of MGMCB. Specifically, we construct a tensor by concatenating multiple contiguous multi-directional  Gaussian derivative kernels along the channel dimension and use them to initialize the weights of one depth-wise convolution. The other depth-wise convolution adopts a 90-degree rotated version of this tensor as its initial weights. To maintain the original multidirectional gradient-convergent properties, the weights of both convolutional layers remain unchanged during training. Finally, the computed $\mathcal{F} _{e}(I_{in} ,k)$ is used to derive the prior components CP1 and CP2.


\textit{3) CP Extraction Modules:} Considering multi-scale object detection, the outputs of MGMCBs at three scales $(3\times3, 5\times5, 7\times7)$ are concatenated and processed through CP extraction modules to generate two representations: a boosted prior map $CP1\in \mathbb{R} ^{W\times H\times 1}$ for precise target localization, and multi-scale hierarchical features $CP2_{i}\in \mathbb{R}  ^{(W/2^{i})\times (H/2^{i})\times(i+1)C} ,i=0,1,2,3$ for subsequent feature embedding, where $C$ denotes the channel dimension. This process can be represented as follows:
\begin{equation}
	\centering
	\begin{aligned}
		CP1=PKE_{1} (\mathcal{F} _{c} (\mathcal{F} _{e}(I_{in},3 ),\mathcal{F} _{e}(I_{in},5), \mathcal{F} _{e}(I_{in},7)))
	\end{aligned}
	\label{eq3_2_4}
\end{equation}
\begin{equation}
	\centering
	\begin{aligned}
		CP2_{i} =PKE_{2} (\mathcal{F} _{c} (\mathcal{F} _{e}(I_{in},3 ),\mathcal{F} _{e}(I_{in},5), \mathcal{F} _{e}(I_{in},7)))
	\end{aligned}
	\label{eq3_2_5}
\end{equation}
where $\mathcal{F} _{c}(\cdot)$ denotes the feature concatenation operation, $PKE_{1} ( \cdot )$  and $PKE_{2} ( \cdot )$  correspond to the extraction processes of the prior components $CP1$ and $CP2_{i}$, respectively, the specific structures of which are shown in Fig. \ref{backbone}. $PKE_{1} ( \cdot )$ employs point-wise convolution and basic operators to generate a single-channel salient map that highlights target regions. $PKE_{2} ( \cdot )$ utilizes a deep architecture with depth-wise separable convolutions and max-pooling modules to construct multiscale feature representations for hierarchical feature fusion. Note that the parameters in $PKE_{1} ( \cdot )$ remain frozen (either averaging or weighted summation) to preserve full-resolution feature extraction, whereas all parameters in $PKE_{2} ( \cdot )$ are kept learnable.

\subsection{DBPEA}
\label{DBPEA}
To dynamically embed CP components to the optimal positions of deep neural networks, we develop DBPEA. The key operational principles of DBPEA include:


\textit{1) CP1 Component Embedding:} 
With the use of CP1's boosted spatial prior encoding target convergence patterns, we first concatenate this structurally informative component with the input image $I_{in}$ along the channel dimension. The concatenated feature map is then processed through a U-shaped architecture, specifically an encoder-decoder subnet, to generate hierarchical features $f_{i}\in \mathbb{R}  ^{(W/2^{i})\times (H/2^{i})\times(i+1)C} ,i=0,1,2,3$.
\begin{equation}
	\centering
	\begin{aligned}
		f_{0},f_{1},f_{2},f_{3} =\mathcal{F} _{x} (\mathcal{F} _{c}(I_{in},CP1 ))
	\end{aligned}
	\label{eq3_3_1}
\end{equation}
where $\mathcal{F} _{x}(\cdot)$ signifies the mapping function learned by the U-shaped architecture. To fully extract the multi-scale feature representation of small targets, we adopt a 4-layer densely nested interactive module (DNIM) \cite{li2022dense}. DNIM enhances feature propagation through dense connections and skip connections bridging encoder-decoder subnets, with the decoder outputs serving as the final hierarchical features.

\textit{2)  CP2 Component Embedding:} To alleviate the degradation of high-level features, we embed the knowledge-encoded CP2 components into the hierarchical features. Inspired by \cite{li2018pyramid} and \cite{hu2018squeeze}, we introduce CHKIM, which utilizes top-down and bottom-up modulation mechanisms  for effective feature fusion.

\begin{figure}
\centering
\subfigure[\footnotesize{Top-Down Attention Modulation}]{\label{subfig_fusion1}
\includegraphics[scale=0.3]{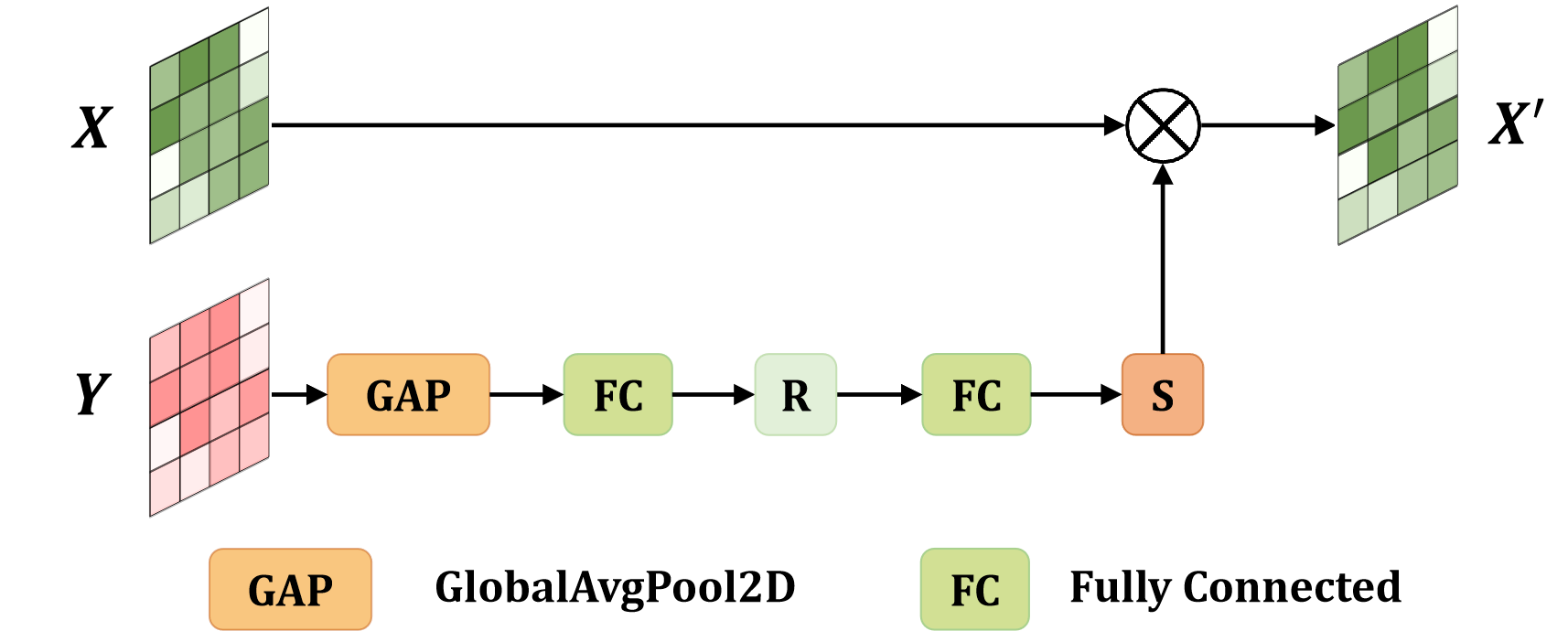}}
\hspace{1cm}
\subfigure[\footnotesize{Bottom-Up Attention Modulation}]{\label{subfig_fusion2}
\includegraphics[scale=0.3]{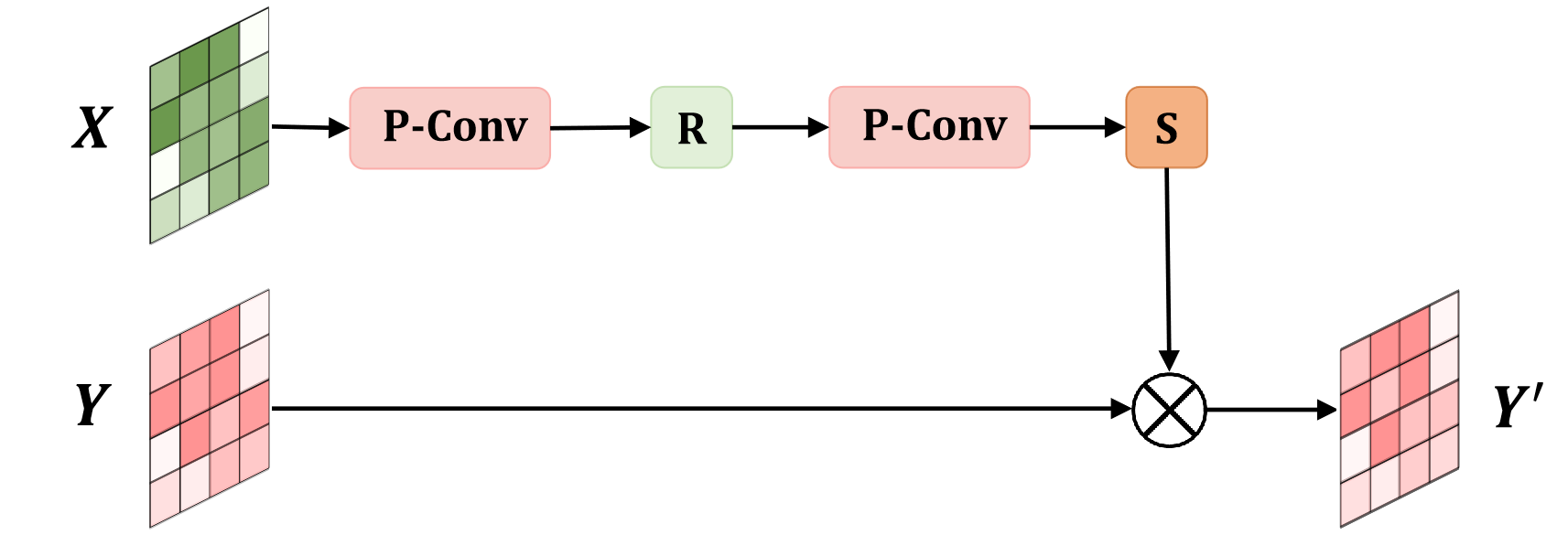}}
\hspace{2cm}
\subfigure[\footnotesize{The Proposed CHKIM}]{\label{subfig_fusion3}
\includegraphics[scale=0.29]{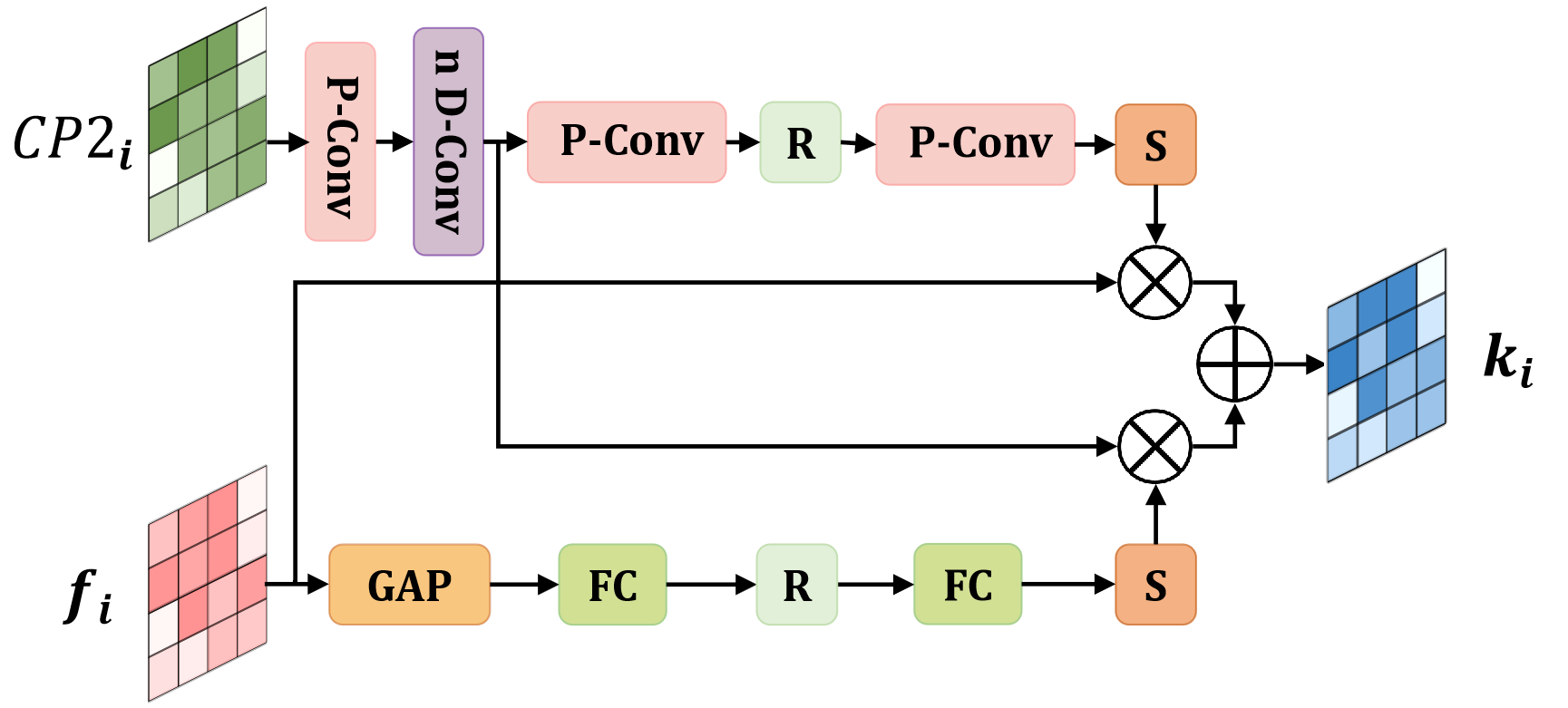}}
   \caption{\small{ Illustration of modulation modules. (a) Top-down global attentional modulation, (b) Bottom-up point-wise attentional modulation, (c) Framework diagram of the proposed CHKIM.\label{fusion_other}}}
\end{figure}

The top-down modulation (shown in Fig. \ref{subfig_fusion1}) suppress redundant feature channels through global high-level semantic information while enhancing discriminative representation. Given a low-level feature $X$ and a high-level feature $Y$ (both with $C$ channels and spatial dimensions $W\times H$), the modulation process is expressed as:
\begin{equation}
	\centering
	\begin{aligned}
		{X}^{\prime}={G}({Y}) \otimes {X}=\sigma\left(\mathcal{B}\left({W}_{2} \delta\left(\mathcal{B}\left({W}_{1} {y}\right)\right)\right)\right) \otimes {X}
	\end{aligned}
	\label{eq3_4_1}
\end{equation}
where $\mathbf{y}=\frac{1}{W \times H} \sum_{i=1, j=1}^{W, H} \mathbf{Y}[:, i, j]$ is the global feature context obtained through global average pooling. $\delta $, $\mathcal{B}$, $\sigma $, and $\otimes $ denote
the rectified linear unit (ReLU) \cite{nair2010rectified}, batch normalization (BN) \cite{ioffe2015batch}, sigmoid function, and element-wise multiplication, respectively.  $W_{1}\in \mathbb{R}^{\frac{C}{r}\times C }$ and $W_{2}\in \mathbb{R}^{C\times \frac{C}{r} }$ are two fully connected layers. $r$ is the channel reduction rate.  

The bottom-up modulation  (shown in Fig. \ref{subfig_fusion2})  enhances the saliency of small targets in deep features through low-level details:
\begin{equation}
	\centering
	\begin{array}{l}
	Y^{'}=L(X)\otimes Y \\ \hspace{3mm} =\sigma\left(\mathcal{B}\left(\mathrm{PWConv}_{2}\left(\delta\left(\mathcal{B}\left(\mathrm{PWConv}_{1}(X)\right)\right)\right)\right)\right)\otimes Y
	\end{array}
	\label{eq3_4_3}
\end{equation}
where $\mathrm{PWConv(\cdot)}$ denotes point-wise convolution \cite{lin2013network}. This convolution operation is commonly used to increase or decrease the number of channels without changing the spatial dimensions of the feature map. The kernel sizes of $\mathrm{PWConv_{1}(\cdot)}$ and $\mathrm{PWConv_{2}(\cdot)}$ are $\frac{C}{4}\times W\times H \times  C$ and $C\times W\times H \times  \frac{C}{4}$, respectively. $L(X)$ has the same shape as $Y$ and can highlight small infrared targets element-wise.

As illustrated in Fig. \ref{subfig_fusion3}, CHKIM fully exploits the advantages of both modulation methods to realize effective feature interaction between shallow CP2 components and deep features $f_{i}$ in an asymmetric manner:
\begin{equation}
	\centering
	\begin{aligned}
	k_{i}=G\left(f_{i}\right) \otimes E\left(CP2_{i}\right)+L(E(CP2_{i})) \otimes f_{i}
	\end{aligned}
	\label{eq3_4_4}
\end{equation}
where $k_{i}\in \mathbb{R}  ^{(W/2^{i})\times (H/2^{i})\times(i+1)C} ,i=0,1,2,3$ is the fused feature, and $E(CP2_{i})=\mathrm{PWConv}(\mathrm{DWConv}(CP2_{i}))$ represents the preprocessing to adjust the spatial scale and channel number of $CP2_{i}$. Through the asymmetric modulation of low-level details and high-level semantics, $CP2_{i}$ and $f_{i}$ can effectively exchange high-level semantics and fine details, achieving richer encoding of semantic information and spatial details.

\begin{figure}[b]
 \centering
 \includegraphics[scale=0.32]{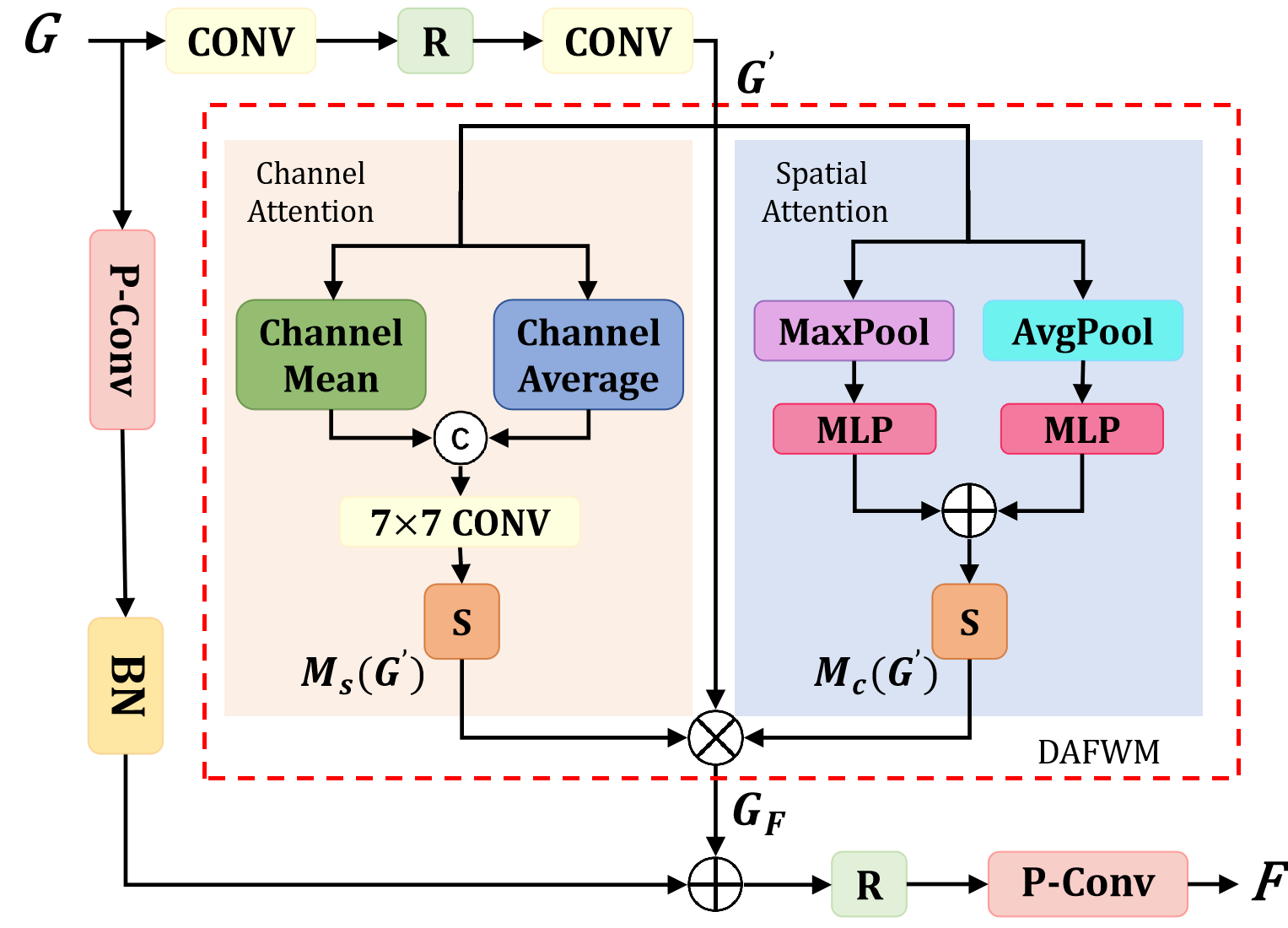}
 \caption{\small{The architecture of AGFEM.}}
 \label{feature_enhance}
\end{figure} 

\subsection{FEM}
\label{The Feature Enhancement Module}
To enhance multi-scale features from multiple layers and improve the utilization rate of features, we design FEM.

First, we adopt a feature pyramid mixing mechanism to aggregate the multi-layer fused features $k_{i} ,i=0,1,2,3$. Specifically, as shown in Fig. \ref{backbone}, each $k_{i} ,i=1,2,3$ is passed through a single $3\times 3$ convolution and up-sampled to the same size, $k_{i}^{up} \in \mathbb{R}  ^{W\times H\times C} ,i=1,2,3$. Subsequently, the features from each layer are concatenated along the channel dimension to obtain a feature map $G=\left \{k_{0},  k_{1}^{up}, k_{2}^{up}, k_{3}^{up}\right \} \in \mathbb{R}  ^{W\times H\times4C} $ with global robustness.

Subsequently, to prevent network degradation, we develop AGFEM to enhance feature representation capabilities through a residual structure, as shown in Fig. \ref{feature_enhance}. Within the feature enhancement backbone, the feature map $G$  is firstly preprocessed through a $3\times3$ convolution and a ReLU activation to obtain $G^{'}\in \mathbb{R}  ^{W\times H\times C} $. Then, the dual-attention feature weighting module (DAFWM) is designed to adaptively enhance features to obtain $G_{F}\in \mathbb{R}  ^{W\times H\times C} $. The DAFWM consists of a channel attention (CA) unit and a spatial attention (SA) unit \cite{wang2017residual}. The CA filters out high-level semantic noise that is irrelevant to the target along the channel dimension, while the SA models global context on the feature map to refine detail features. Weighting feature maps with channel and spatial attention can enhance high-resolution detail and semantic abstraction. The DAFWM process is given by:
\begin{equation}
	\centering
	\begin{array}{l}
		M_{c}\left(G^{\prime}\right)=\sigma\left(M L P\left(P_{\max}\left(G^{\prime}\right)\right)+M L P\left(P_{\text {avg}}\left(G^{\prime}\right)\right)\right)\\[5pt]
M_{s}\left(G^{\prime}\right)=\sigma\left(\operatorname{Conv}^{7 \times 7}\left( \mathcal{F} _{c} \left(CP_{\max}\left(G^{\prime}\right), CP_{\text {avg}}\left(G^{\prime}\right)\right)\right)\right) \\[5pt]
 G_{F}=\left(M_{c}\left(G^{\prime}\right) \otimes M_{s}\left(G^{\prime}\right)\right) \otimes G^{\prime} 
	\end{array}
	\label{eq3_5_1}
\end{equation}
where  $M_{c}(G^{'})$ and  $M_{s}(G^{'})$ represent the channel attention map and spatial attention map, respectively. $P_{\max}(\cdot)$ and $P_{\text {avg}}(\cdot)$ denote the adaptive maximum pooling layer and the adaptive average pooling layer, respectively. The shared network is composed of a multi-layer perceptron (MLP) with one hidden layer. $\operatorname{Conv}^{7 \times 7}(\cdot)$  represents the convolution operation with a $7 \times 7$ filter kernel.  $CP_{\max}(\cdot)$ and $CP_{\text {avg}}(\cdot)$ denote the operations of computing the maximum and average values along the channels, respectively. 

Finally, to ensure that the gradients of DAFWM do not excessively vanish or explode during backpropagation, and to preserve potential target information in the original features to avoid over-suppression of non-salient regions by the attention mechanism, we add  $G$ and $G_{F}$ using a residual structure. The feature map $G$ is first passed through a point-wise convolution and BN to match the scale with $G_{F}$, then added element-wise to the attention-weighted feature $G_{F}$, followed by ReLU activation and point-wise convolution to obtain the final feature map $F\in\mathbb{R}^{W\times H\times 1}$.
\begin{equation}
	\centering
	\begin{aligned}
	F=\mathrm {PWConv}(\delta (\mathcal{B}(\mathrm {PWConv}(G))\oplus  G_{F})
	\end{aligned}
	\label{eq3_5_2}
\end{equation}

\section{Experiments And Analysis}

\subsection{Datasets and Evaluation Metrics}
\textit{1) Datasets:} We use three publicly available datasets, including NUDT-SIRST \cite{li2022dense}, IRSTD-1k \cite{zhang2022isnet} and NUAA-SIRST \cite{dai2021asymmetric}. The NUDT-SIRST dataset consists of 1327 images, with 663 images for training and 664 for testing, and images are sized at 256 × 256 pixels. This dataset covers five major scenes: urban, rural, ocean, heavy cloud, and highlight. The IRSTD-1K dataset contains 1001 images, with 800 for training and 201 for testing, and the image size is 512 × 512 pixels. It includes targets of various shapes and sizes with cluttered backgrounds. The NUAA-SIRST dataset comprises 427 images, with 213 for training and 214 for testing. The targets are faint and hidden in complex background environments, and are also subject to interference from strong light sources, sheet-like clouds, and sea surfaces.

\textit{2) Evaluation Metrics:} We compare CSPENet with the SOTA methods using several common metrics, including intersection over union $(IoU)$, F-measure $(F_{1})$, probability of detection $(P_{d})$, and false-alarm rate $(F_{a})$.

$IoU$ is used to measure the degree of overlap between the predicted target region by the model and the ground-truth target region. It is described as:
\begin{equation}
	\centering
	\begin{array}{l}
		IoU=\dfrac{\text { Area of Overlap }}{\text { Area of Union }}
	\end{array}
	\label{eq100}
\end{equation}

$F_{1}$ measures precision and recall, where precision represents the ratio of correctly predicted target pixels to the total predicted target pixels, and recall represents the ratio of correctly predicted target pixels to the total ground-truth target pixels. Its definition formula is as follows:
\begin{equation}
	\centering
	\begin{array}{l}
		Precision =\dfrac{T P}{T P+F P} \\[15pt]
Recall =\dfrac{T P}{T P+F N} \\[15pt]
F_{1}=\dfrac{2 * Precision  *  Recall }{ Precision + Recall }
	\end{array}
	\label{eq101}
\end{equation}

$P_{d}$ reflects the ability to correctly detect targets, calculated as the ratio of correctly detected targets to the total number of actual targets. Its expression is given as follows:
\begin{equation}
	\centering
	\begin{array}{l}
		P_{d}=\dfrac{T_{correct}}{T_{all}} 
	\end{array}
	\label{eq102}
\end{equation}

$F_{a}$ reflects the accuracy of target detection, defined as the ratio of the number  of incorrectly predicted pixels to the total number of pixels in the entire image. Its formula is expressed as:
\begin{equation}
	\centering
	\begin{array}{l}
		F_{a}=\dfrac{P_{false}}{P_{all}} 
	\end{array}
	\label{eq103}
\end{equation}

\subsection{Implementation Details}
Our model is implemented using the PyTorch framework on a Nvidia GeForce RTX 4090 GPU. During network training, the batch size is set to 8, and the number of epochs is set to 400. We employ the Adam optimizer \cite{kingma2014adam}  with an initial learning rate of $5\times 10^{-6} $, and use MultiStepLR to dynamically adjust the learning rate.

\subsection{Comparison to SOTA Methods}
To demonstrate the superiority of our method, we compare our CSPENet to several SOTA methods, including model-driven methods (MPCM \cite{wei2016multiscale}, PSTNN \cite{zhang2019infrared}, IPI \cite{gao2013infrared}) and DL-based methods (ACM  \cite{dai2021asymmetric}, ALCNet \cite{dai2021attentional}, AGPCNet \cite{zhang2021agpcnet}, DNANet \cite{li2022dense}, UIUNet \cite{wu2022uiu}, RDIAN \cite{sun2023receptive}, RPCANet \cite{wu2024rpcanet}, MDIGCNet \cite{zhang2024mdigcnet}) on the three public datasets: NUDT-SIRST \cite{li2022dense}, IRSTD-1k \cite{zhang2022isnet}, and NUAA-SIRST \cite{dai2021asymmetric}.

\begin{table*}[htbp!]
    \centering
    \scriptsize
    \caption{\\\footnotesize \scshape Comparison \scshape Of \scshape Detection \scshape Performance[$IoU\left ( \% \right ) $, $F_{1}\left ( \% \right ) $, $P_{d}\left ( \% \right ) $, AND $F_{a}\left ( \times 10^{-6}  \right ) $] \scshape And \scshape Model \scshape Efficiency [\scshape The \scshape Number \scshape Of \scshape Parameters (M) \scshape And \scshape Theoretical \scshape Flops (G)] \scshape Of \scshape Different \scshape Methods \scshape On \scshape Nudt-Sirst, \scshape Irstd-1k, \scshape And \scshape Nuaa-Sirst. \scshape The \scshape Best \scshape Results \scshape Are \scshape In \scshape Red, \scshape And \scshape The \scshape Second \scshape Best \scshape Results \scshape Are \scshape In \scshape Blue.}
    \label{all_result}
     \begin{tabular*}{\textwidth}{@{\extracolsep\fill}l|c|c|cccc|cccc|cccc}
    \hline
     \multirow{2}{*}{ Methods} &  \multirow{2}{*}{Params(M) } &  \multirow{2}{*}{FLOPs(G) } & \multicolumn{4}{c|}{ NUDT-SIRST \cite{li2022dense}}& \multicolumn{4}{c|}{IRSTD-1K \cite{zhang2022isnet} }& \multicolumn{4}{c}{NUAA-SIRST \cite{dai2021asymmetric} } \\\cline{4-15}
      && & $IoU$↑  & $F_{1}$↑& $P_{d}$↑& $F_{a}$↓& $IoU$↑ & $F_{1}$↑& $P_{d}$↑& $F_{a}$↓& $IoU$↑ & $F_{1}$↑& $P_{d}$↑& $F_{a}$↓ \\\hline
       MPCM \cite{wei2016multiscale} &- &-   &8.18  &15.12 &66.14  &115.96  &7.33 &25.89  &68.73  &65.12 &14.87 &36.27 &80.06 &51.43 \\
       PSTNN \cite{zhang2019infrared} &- &- &21.60   &40.91  &70.83 &36.19  &15.94  &28.63 &67.83  &83.71  &22.35 &24.31 &77.95 &54.43 \\
       IPI \cite{gao2013infrared}&-  &- &27.60   &31.82  &72.28 &37.28    &27.60 &31.82 &72.28 &37.28  &25.67  &40.18 &83.27  &38.92\\\hline
       ACM \cite{dai2021asymmetric}&0.398 &0.40 &68.47   &79.06  &95.18 &17.58  &64.17  &78.82 &90.57  &92.64  &69.76 &83.28 &91.89 &28.51 \\
       ALCNet \cite{dai2021attentional}&0.378 &3.74 &81.24    &89.95  &97.57 &14.37  &66.02  &79.06 &92.58  &69.45  &70.21 &83.46 &92.78 &40.34 \\
       AGPCNet \cite{zhang2021agpcnet}&12.362 &43.18 &87.94   &90.58  &97.49 &7.92  &66.19  &79.35 &89.53  &26.61  &75.16 &85.32 &93.91 &45.27 \\
       DNANet \cite{li2022dense}&4.697 &14.26 &92.79   &{\color{blue}96.26}  &{\color{red}98.52} &{\color{red}4.52}  &65.23  &79.75 &{\color{blue}92.93}  &40.04  &74.03 &{\color{red}87.79} &{\color{blue}95.82} &43.36 \\
       UIUNet \cite{wu2022uiu}&50.540 &54.42 &90.39   &95.92  &97.57 &7.91  &66.20  &78.82 &90.91  &{\color{red}20.21}  &73.48 &87.01 &92.40 &{\color{blue}18.99}\\
        RDIAN \cite{sun2023receptive}&0.022 &3.72 &85.70   &92.18  &{\color{blue}97.88} &22.81  &66.02  &79.69 &91.57  &36.44  &69.81 &84.88 &94.42 &58.58 \\
       RPCANet \cite{wu2024rpcanet}&0.680 &1.96 &89.31   &94.35  &97.14 &28.73  &{\color{blue}66.93}  &{\color{blue}80.59} &89.35  &43.90  &74.21 &86.51 &95.62 &47.21 \\
       MDIGCNet \cite{zhang2024mdigcnet}&1.505 &6.557 &{\color{blue}93.41}  &96.03  &96.67 &37.28  &{\color{red}67.09}  &80.44 &91.25  &40.38  &{\color{blue}76.23} &82.37 &93.65 &56.84 \\
       \textbf{Ours}&1.440 &9.49 &{\color{red}94.18}  &{\color{red}97.00}  &{\color{red}98.52} &{\color{blue}4.92}  &66.79  &{\color{red}81.10} &{\color{red}93.27}  &{\color{blue}23.53}  &{\color{red}79.83} &{\color{blue}87.41} &{\color{red}96.96} &{\color{red}15.18} \\\hline
    \end{tabular*}
\end{table*}

\begin{figure*}
\centering
\subfigure[]{\label{ROC_1}
\includegraphics[scale=0.45]{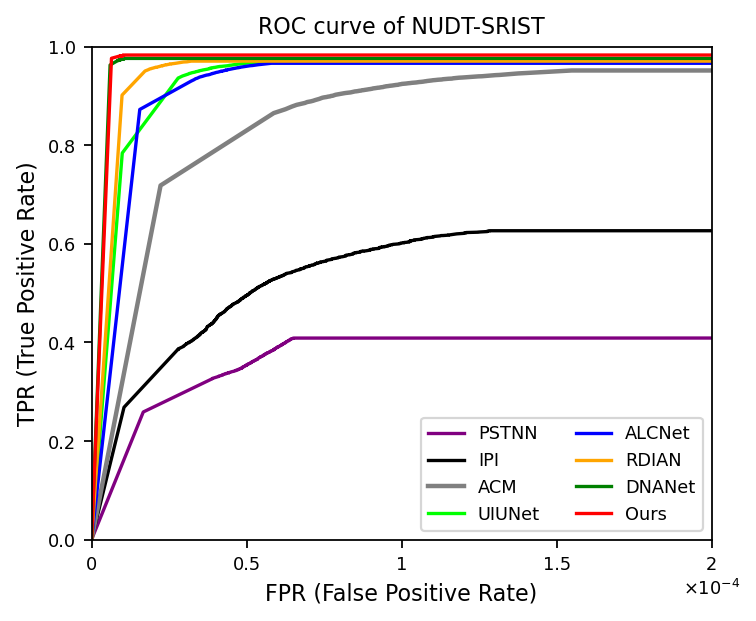}}
\hspace{0.5cm}
\subfigure[]{\label{ROC_2}
\includegraphics[scale=0.45]{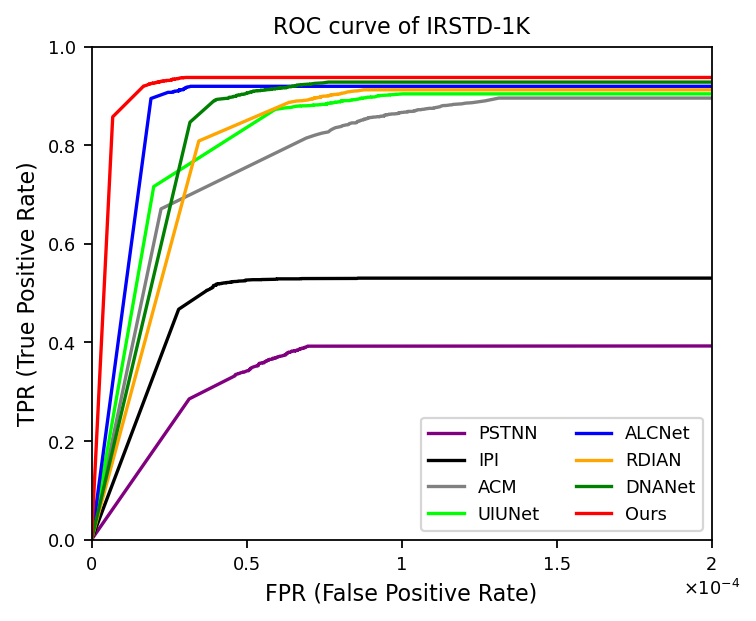}}
\hspace{0.5cm}
\subfigure[]{\label{ROC_3}
\includegraphics[scale=0.45]{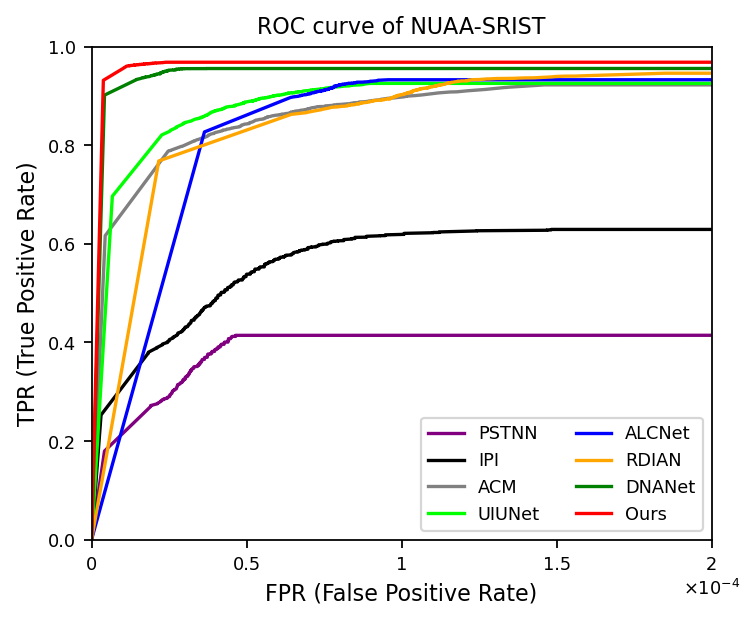}}
   \caption{ \small{ROC curves of different algorithms. (a) ROC curve of different algorithms on NUDT-SIRST. (b) ROC curve of different algorithms on IRSTD-1K. (c) ROC curve of different algorithms on NUAA-SIRST.}\label{ROC}}
\end{figure*}

\textit{1) Quantitative Results:}
The quantitative comparison of these algorithms is presented in Table \ref{all_result}. The best results are in red, and the second best results are in blue.  The symbols ↑ and ↓ denote metrics
where higher and lower values are preferable, respectively. Specifically, although model-driven methods such as MPCM, PSTNN and IPI  have relatively high $P_{d}$, they perform worse on the pixel-level metrics $IoU$ and $F_{1}$ compared to DL-based methods, indicating insufficient detail preservation. ACM and AGPCNet demonstrate significant improvements over model-driven methods, but their performance is generally average among DL algorithms. ALCNet, despite having a relatively good $P_{d}$, only achieves an $F_{1}$ of 89.95$\%$ on NUDT-SIRST, while others exceed 90.00$\%$. MDIGCNet achieves superior $IoU$ scores across the three datasets, but its results in $P_{d}$ are not particularly outstanding. In contrast, DNANet shows overall good performance, especially achieving the highest $P_{d}$ of 98.52$\%$ on NUDT-SIRST.

The proposed CSPENet demonstrates competitive performance
with relatively low computational overhead. Specifically, CSPENet achieves the preeminent $P_{d}$ on three datasets, with especially pronounced advantages on IRSTD-1K and NUAA-SIRST, where it outperforms the second-ranking algorithms by a margin of 0.34$\%$ and 1.14$\%$, respectively. Compared to the parameter-efficient RDIAN and ALCNet, our CSPENet demonstrates superior performance across all metrics on the three datasets. Specifically, on NUDT-SIRST, our method realizes a remarkable $IoU$ that is 12.94$\%$ higher, an $F_{1}$ that is 7.05$\%$ elevated, a $P_{d}$ that is 0.95$\%$ augmented, and a $F_{a}$ that is reduced by 9.45 × 10$^{-5} $ in comparison to ALCNet. In comparison with MDIGCNet, which has slightly higher parameters and computational costs, our method achieves substantial enhancements in $P_{d}$, with increases of 1.85$\%$, 2.02$\%$, and 3.31$\%$ on the three datasets, respectively. Other metrics are comparable to or better than those of MDIGCNet. Compared with DNANet, AGPCNet, and UIUNet, which require more parameters and higher computational costs, our CSPENet consistently realizes superior detection performance across all three datasets with less parameters. This indicates that our method balances performance and computational overhead, offering greater practicality.

Furthermore, we plot the ROC curves of various methods on NUDT-SIRST, IRSTD-1k, and NUAA-SIRST. As shown in Fig. \ref{ROC}, our CSPENet rapidly approaches the top-left corner across all datasets, surpassing all traditional algorithms and nearly all DL algorithms. For example, in the ROC curve of IRSTD-1K, our method reaches the corner first. Although algorithms such as DNANet and UIUNet exhibit relatively high $P_{d}$ or low $F_{a}$ in Table \ref{all_result}, they cannot compete with our method in the early stages of lower thresholds. This indicates that our method possesses exceptional detection performance with fewer false alarms.



\begin{figure*}
 \centering
 \includegraphics[width=1\linewidth]{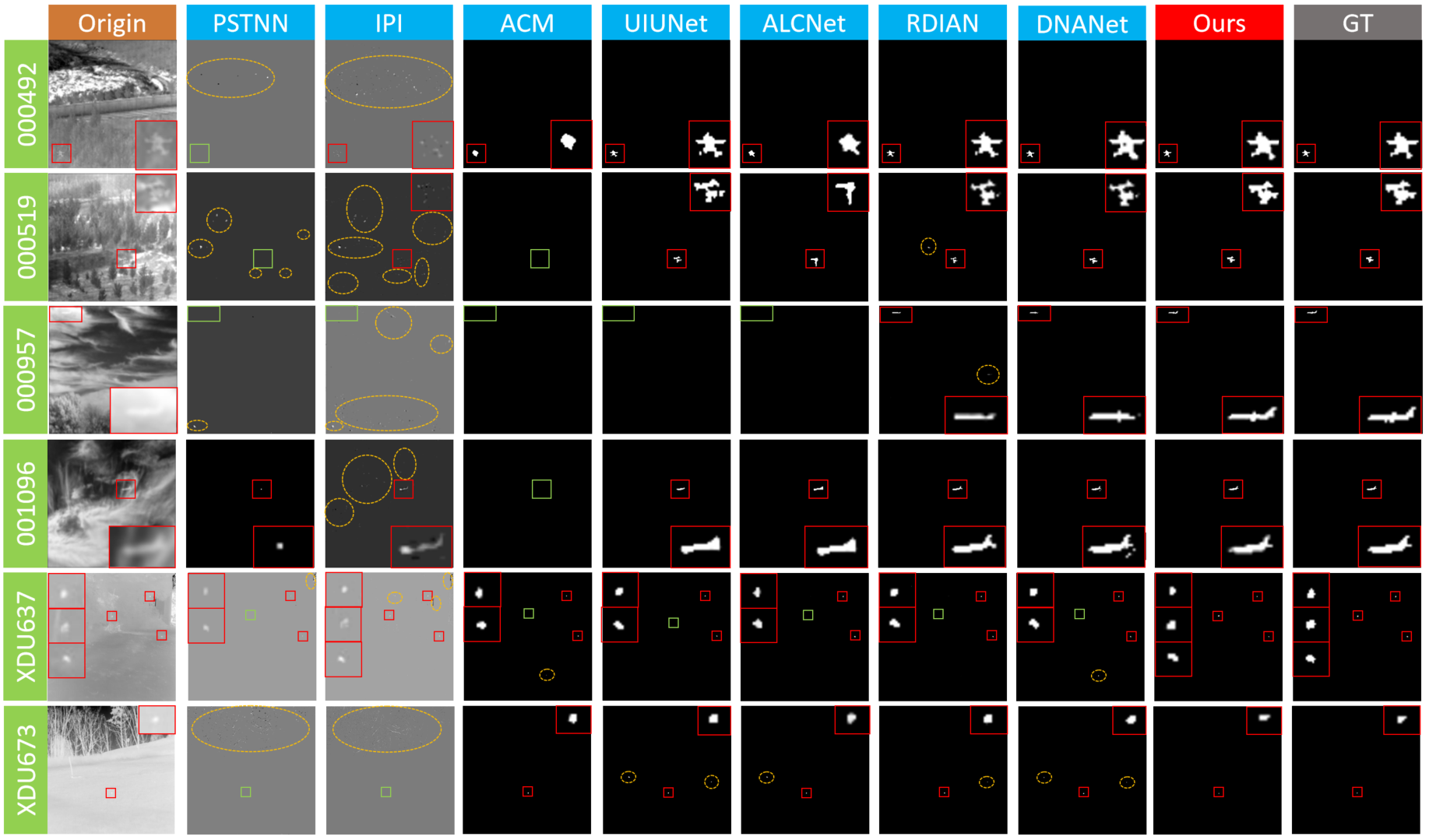}
 \caption{\small{Visual examples of different methods on three different datasets, with yellow circles indicating false positives and green boxes indicating missed detections. The detected target areas are highlighted in red and enlarged, placed in the corners of the detection image.}}
 \label{result_img}
 \vspace{-0.5cm}
\end{figure*}

\begin{table*}[htbp!]
    \centering
    \scriptsize
    \vspace{0.5cm}
    \caption{\\\footnotesize \scshape Ablation \scshape Experiments \scshape of  \scshape Quantitative \scshape Metrics $[$$IoU\left ( \% \right ) $, $F_{1}\left ( \% \right ) $, $P_{d}\left ( \% \right ) $, AND $F_{a}\left ( \times 10^{-6}  \right ) $  $]$  \scshape With \scshape Different \scshape Prior \scshape Knowledge \scshape Embedded \scshape Into \scshape The \scshape Backbone \scshape Network. \scshape The \scshape Best \scshape Results \scshape Are \scshape In \scshape Red, \scshape And \scshape The \scshape Second \scshape Best \scshape Results \scshape Are \scshape In \scshape Blue.}
    \label{Ablation_SCPEM_1}
     \begin{tabular*}{\textwidth}{@{\extracolsep\fill}l|cccc|cccc|cccc}
    \hline
     \multirow{2}{*}{ Methods}  & \multicolumn{4}{c|}{ NUDT-SIRST \cite{li2022dense}}& \multicolumn{4}{c|}{IRSTD-1K \cite{zhang2022isnet} }& \multicolumn{4}{c}{NUAA-SIRST \cite{dai2021asymmetric} } \\\cline{2-13}
        & $IoU$↑  & $F_{1}$↑& $P_{d}$↑& $F_{a}$↑& $IoU$↑ & $F_{1}$↑& $P_{d}$↑& $F_{a}$↑& $IoU$↑ & $F_{1}$↑& $P_{d}$↑& $F_{a}$↑ \\\hline
       backbone + Top-hat \cite{bai2010infrared} &85.70	&91.58	&96.55	&17.85	&64.23	&79.69	&90.91	&40.04	&74.03 	&85.23	&95.82	&40.34 \\
       backbone + LCM \cite{chen2013local} &87.26	&92.04	&97.57	&18.54	&64.39	&79.98	&91.25	&42.56	&74.21	&85.45	&96.20	&45.86 \\
       backbone + MPCM \cite{wei2016multiscale} &85.44	&92.18	&96.98	&23.76	&64.78	&79.75	&91.57	&48.30	&74.58	&85.39	&95.82	&47.21 \\
      backbone + IPI \cite{gao2013infrared} &90.79	&{\color{blue}94.35}	&97.57	&7.82	&{\color{blue}65.23}	&80.44	&{\color{blue}92.93}	&39.52	&{\color{blue}75.16}	&86.51	&{\color{blue}96.58}	&38.92 \\
      backbone + PSTNN \cite{zhang2019infrared}  &{\color{blue}92.34}	&93.95	&{\color{blue}97.99}	&{\color{blue}6.31}	&65.17	&{\color{blue}80.59}	&{\color{blue}92.93}	&{\color{blue}36.44}	&74.92	&{\color{blue}87.01}	&96.41	&{\color{blue}36.20}\\\hline
     \textbf{backbone + SCPEM}& {\color{red}94.18}  & {\color{red}97.00}  & {\color{red}98.52} &{\color{red}4.92}  &{\color{red}66.79}  &{\color{red}81.10} &{\color{red}93.27}  &{\color{red}23.53}  &{\color{red}79.83} &{\color{red}87.41} &{\color{red}96.96} &{\color{red}15.18} \\\hline
    \end{tabular*}
    
\end{table*}

\textit{2) Visual Comparison:} We present the visualization outcomes of compared methods under challenging scenarios across three datasets in Fig. \ref{result_img}. In each figure, yellow circles indicate false positives, green boxes indicate missed detections, and the detected target areas are highlighted in red and enlarged, placed in the corner of the detection image. It is worth noting that traditional methods, exemplified by IPI and PSTNN, often generate a significant number of false positives, especially in cases where the background contains strong edges, as shown in images 000519 from NUDT-SIRST and XDU673 from IRSTD-1K. Moreover, the shapes of the targets detected by traditional methods significantly differ from the ground truth (GT), making it difficult to accurately reconstruct the actual shape of the targets, sometimes even misidentifying a single target as multiple targets.
DL-based methods, such as ACM and UIUNet, also exhibit missed detections in high-illumination scenes, for instance, in images 000957 from NUDT-SIRST and XDU637 from IRSTD-1K. Additionally, these methods are prone to erroneously classifying parts of irregularly shaped targets as background, resulting in suboptimal performance in delineating target contours and an inability to accurately detect the complete shape of the target, as seen in images 000492, 000519, and 001096 from NUDT-SIRST.

In contrast, our CSPENet demonstrates excellent performance in multi-target and high-brightness scenarios, accurately detecting all targets with the lowest false alarm rate. In addition, the target contour is depicted more accurately. By embedding the two types of priors obtained from SCPEM via DBPEA into the network, CSPENet is able to locate small targets more accurately and extract richer contour details of target-like objects. Furthermore, the use of CHKIM for multi-hierarchical feature fusion and AGFEM for feature enhancement improves feature utilization efficiency, allowing the detection of dim targets against similar gray backgrounds, thereby reducing the occurrence of missed detections. 

\subsection{Ablation Study}
To illustrate the effectiveness of the proposed CSPENet, we perform ablation studies on the key modules and evaluate their performance on the aforementioned three datasets. The structure of the rest of the network remains unchanged throughout the experiments.

\textit{1) Effect of SCPEM:} SCPEM is designed to extract CP knowledge, generating two collaborative CP components that possess both spatial structural information and contour detail retention capabilities. To investigate the benefits brought by the prior knowledge obtained by this module, we first compare the results of using the original infrared image processed by SCPEM with those from other methods (such as Top-hat, LCM, MPCM, IPI, and PSTNN) as structural priors embedded into the backbone network through DBPEA, as shown in Table \ref{Ablation_SCPEM_1}. Among them, Top-hat performs filtering on the image, LCM and MPCM enhance the contrast of the image, and IPI and PSTNN provide sparse results of the image.  It can be observed that on NUDT-SIRST, IRSTD-1K and NUAA-SIRST, the $IoU$, $F_{1}$, $P_{d}$ and $F_{a}$ all reach the optimal values. Specifically, $IoU$ improves by at least 1.84$\%$, 1.56$\%$ and 4.67$\%$, $F_{1}$ improves by at least 2.65$\%$, 0.51$\%$ and 0.3$\%$, $P_{d}$  improves by at least 0.53$\%$, 0.34$\%$ and 0.38$\%$, and $F_{a}$ reduces by at least 1.39$\times$10$^{-6}$, 12.91$\times$10$^{-6}$ and 21.02$\times$10$^{-6}$.

\begin{figure*}
 \centering
 \includegraphics[width=1\linewidth]{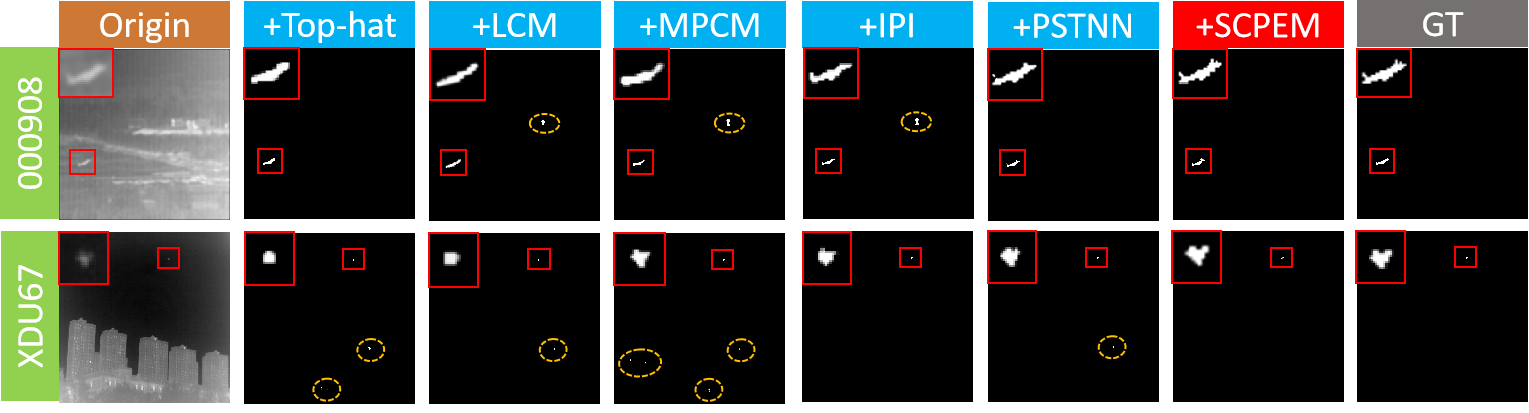}
 \caption{\small{Visual examples of embedding different prior knowledge into the backbone on three different datasets, with yellow circles indicating false positives. The detected target areas are highlighted in red and enlarged, placed in the corners of the detection image.}}
 \label{abl_PK_others}
\end{figure*}

Meanwhile, we present the visualization results of different methods in challenging scenarios on three datasets in Fig. \ref{abl_PK_others}. It can be seen that using the results of SCPEM as prior knowledge embedded into deep learning networks can accurately locate target positions, eliminate background clustering, and preserve key contour features, which proves the effectiveness of this module.

\begin{figure*}
 \centering
 \includegraphics[width=1\linewidth]{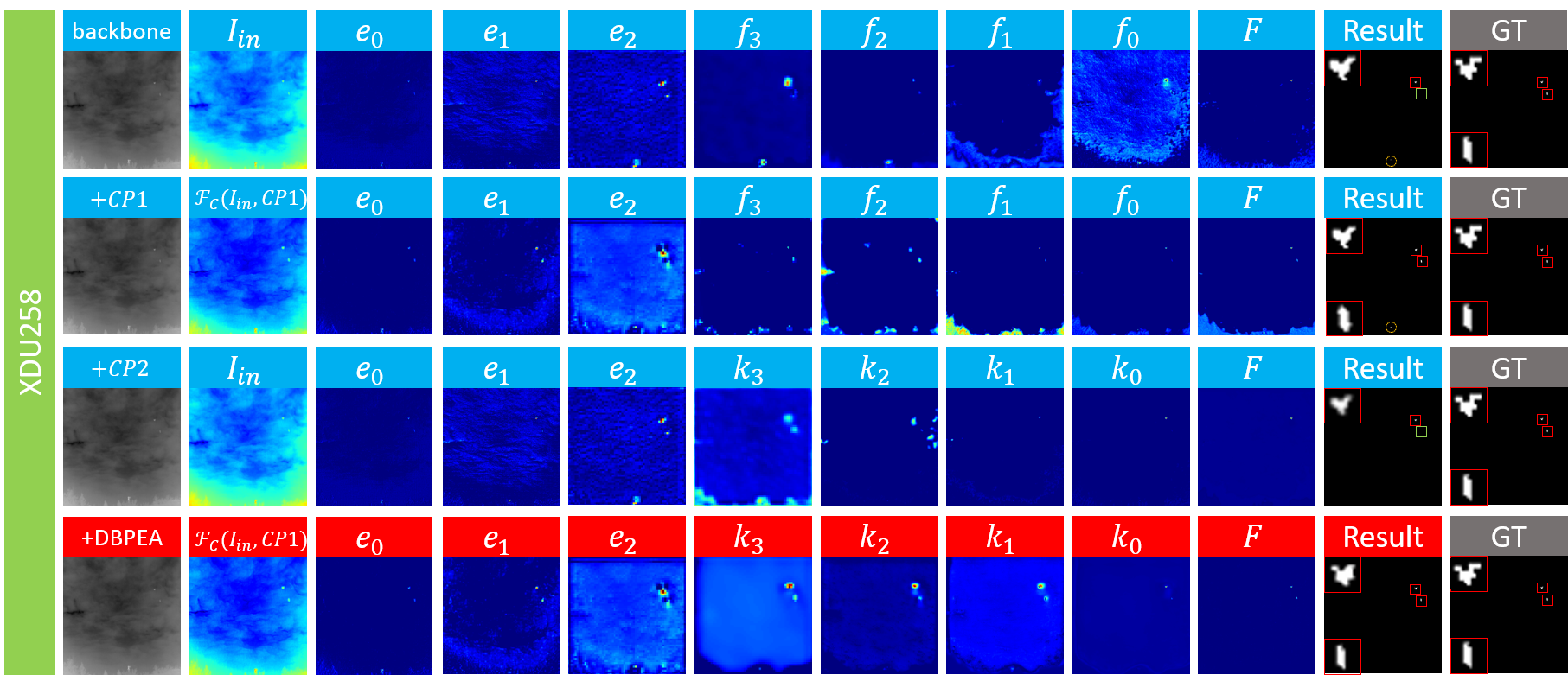}
 \caption{\small{Heatmap and description of detection results for an image with severe background clutter interference. The first to fourth rows of pictures represent backbone, backbone+CP1 , backbone+CP2 and backbone+DBPEA, respectively.}}
 \label{PK_feature_1}
\end{figure*}

\begin{table*}[htbp!]
    \centering
    \scriptsize
    \caption{\\\footnotesize $IoU\left ( \% \right ) $, $F_{1}\left ( \% \right ) $, $P_{d}\left ( \% \right ) $, AND $F_{a}\left ( \times 10^{-6}  \right ) $\scshape Values \scshape Achieved \scshape In \scshape Nudt-Sirst, \scshape Irstd-1k, \scshape And \scshape Nuaa-Sirst \scshape On \scshape Ablation \scshape Experiments \scshape About \scshape Dbpea. \scshape The \scshape Best \scshape Results \scshape Are \scshape In \scshape Red, \scshape And \scshape The \scshape Second \scshape Best \scshape Results \scshape Are \scshape In \scshape Blue.
}
    \label{Ablation_DBPEA_1}
     \begin{tabular*}{\textwidth}{@{\extracolsep\fill}cc|cccc|cccc|cccc}
    \hline
     \multirow{2}{*}{ $CP1$} &\multirow{2}{*}{ $CP2$}  & \multicolumn{4}{c|}{ NUDT-SIRST \cite{li2022dense}}& \multicolumn{4}{c|}{IRSTD-1K \cite{zhang2022isnet} }& \multicolumn{4}{c}{NUAA-SIRST \cite{dai2021asymmetric} } \\\cline{3-14}
       & & $IoU$↑  & $F_{1}$↑& $P_{d}$↑& $F_{a}$↓& $IoU$↑ & $F_{1}$↑& $P_{d}$↑& $F_{a}$↑& $IoU$↑ & $F_{1}$↑& $P_{d}$↑& $F_{a}$↑ \\\hline
       \textbf{ $\times $ } &\textbf{ $\times $ } &90.39   &95.92  &97.57 &7.91  &65.20  &78.82 &90.91  &40.04  &73.48 &{\color{blue}87.01} &92.40 &43.36 \\
      \textbf{ $\surd$ }  &\textbf{ $\times $ }&{\color{blue}94.15}	&{\color{blue}96.98}	&{\color{blue}98.41}	&{\color{red}4.00}	&{\color{blue}66.29}	&{\color{blue}80.68}	&90.57	&{\color{blue}26.27}	&{\color{blue}74.51}	&85.39	&{\color{red}96.96}	&{\color{blue}36.20} \\
       \textbf{ $\times $ }  &\textbf{ $\surd$ }  &93.34	&96.55	&{\color{blue}98.41}	&6.60	&64.39	&78.94	&{\color{blue}92.26}	&36.91	&73.16	&84.50	&{\color{blue}96.20}	&46.01\\\hline
     \textbf{ $\surd$ }  &\textbf{ $\surd$ }   & {\color{red}94.18}  & {\color{red}97.00}  & {\color{red}98.52} &{\color{blue}4.92}  &{\color{red}66.79}  &{\color{red}81.10} &{\color{red}93.27}  &{\color{red}23.53}  &{\color{red}79.83} &{\color{red}87.41} &{\color{red}96.96} &{\color{red}15.18} \\\hline
    \end{tabular*}
\end{table*}

\begin{figure*}
 \centering
 \includegraphics[width=1\linewidth]{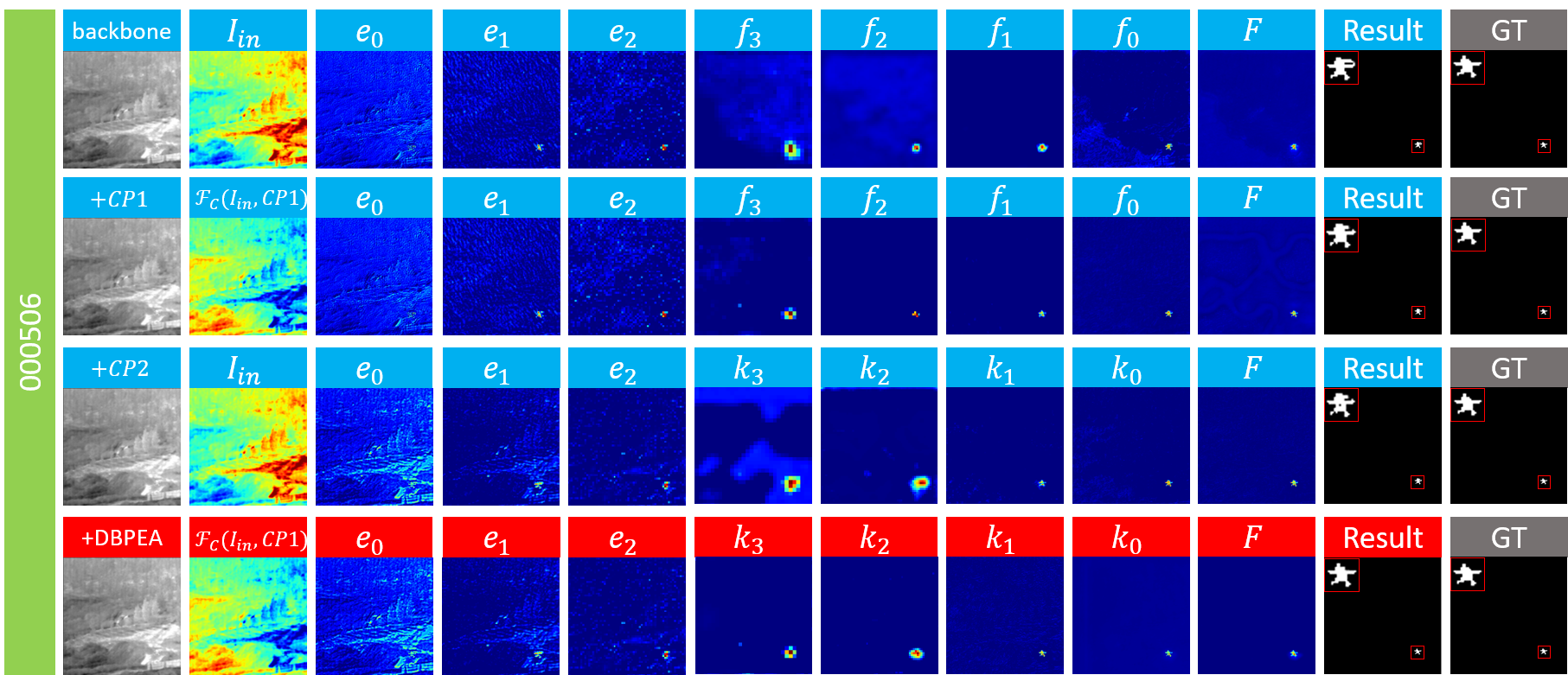}
 \caption{\small{Heatmap and description of detection results for an image containing an irregularly shaped target. The first to fourth rows of pictures represent backbone, backbone+CP1 , backbone+CP2  and backbone+DBPEA respectively.}}
 \label{PK_feature_2}
\end{figure*} 

\begin{figure*}
\centering
\subfigure[\small{Local}]{\label{subfig_abs_fusion1}
\includegraphics[scale=0.23]{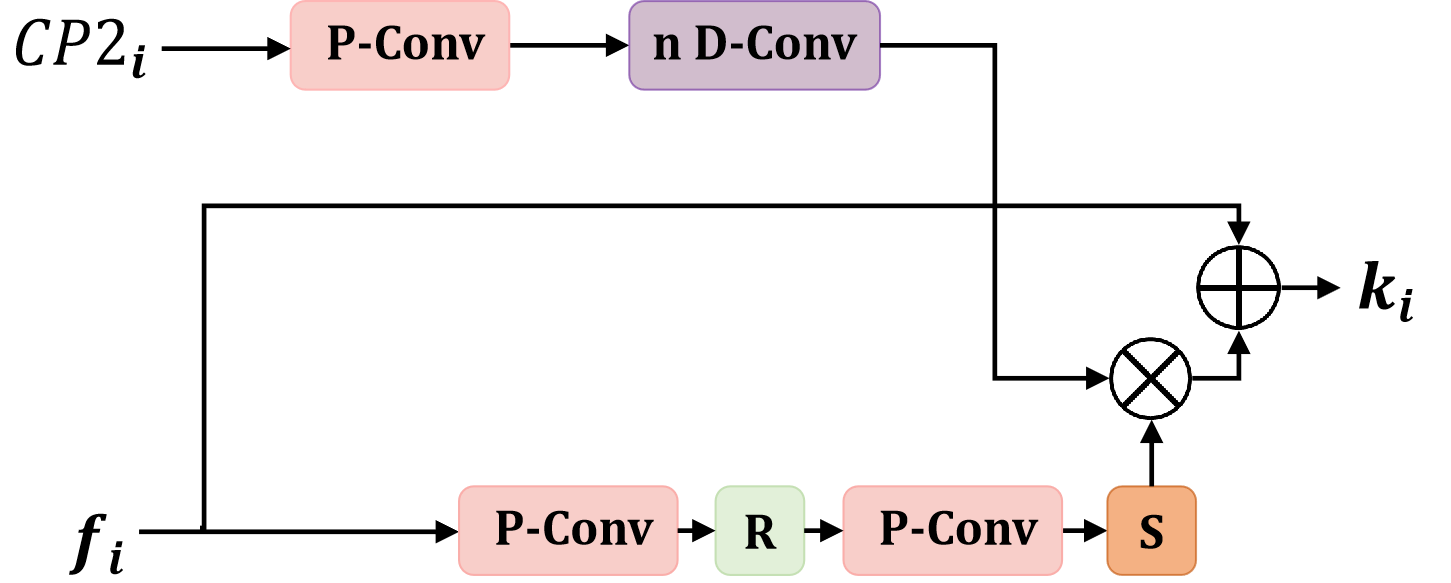}}
\hspace{0cm}
\subfigure[\small{BiLocal}]{\label{subfig_abs_fusion2}
\includegraphics[scale=0.23]{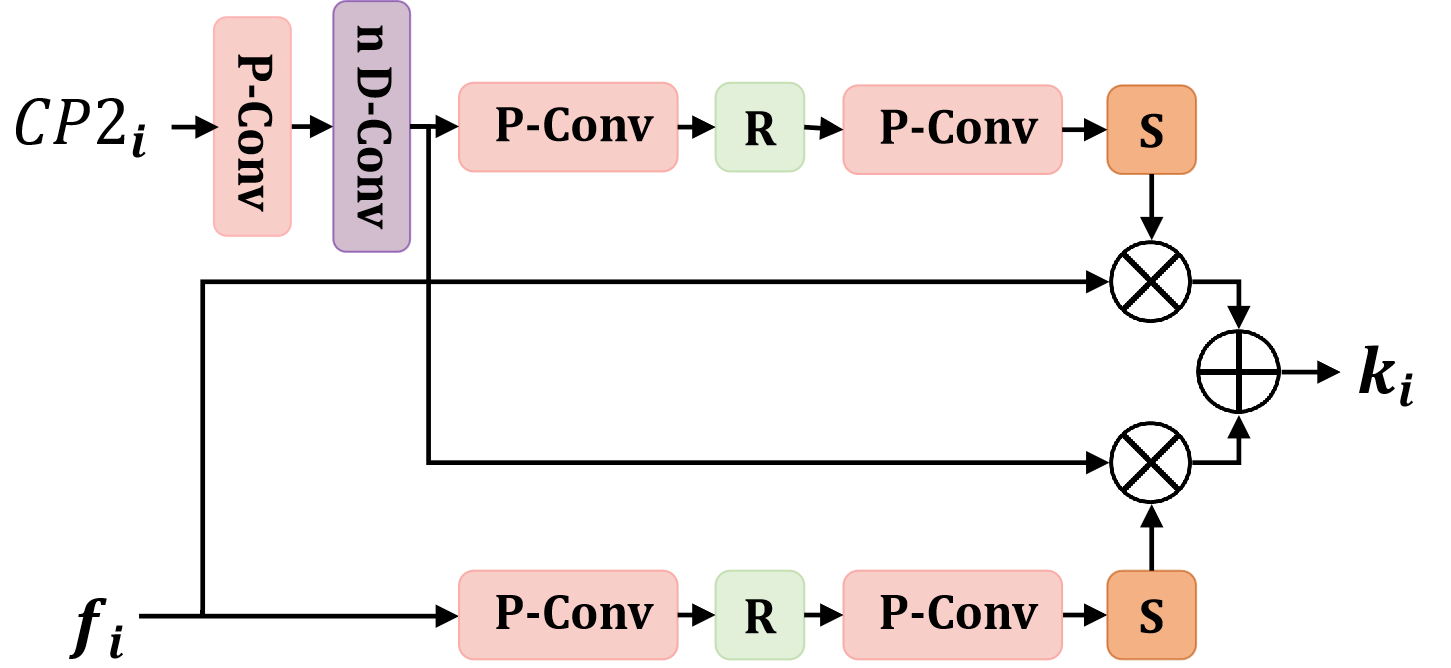}}
\hspace{0cm}
\subfigure[\small{BiGlobal}]{\label{subfig_abs_fusion3}
\includegraphics[scale=0.23]{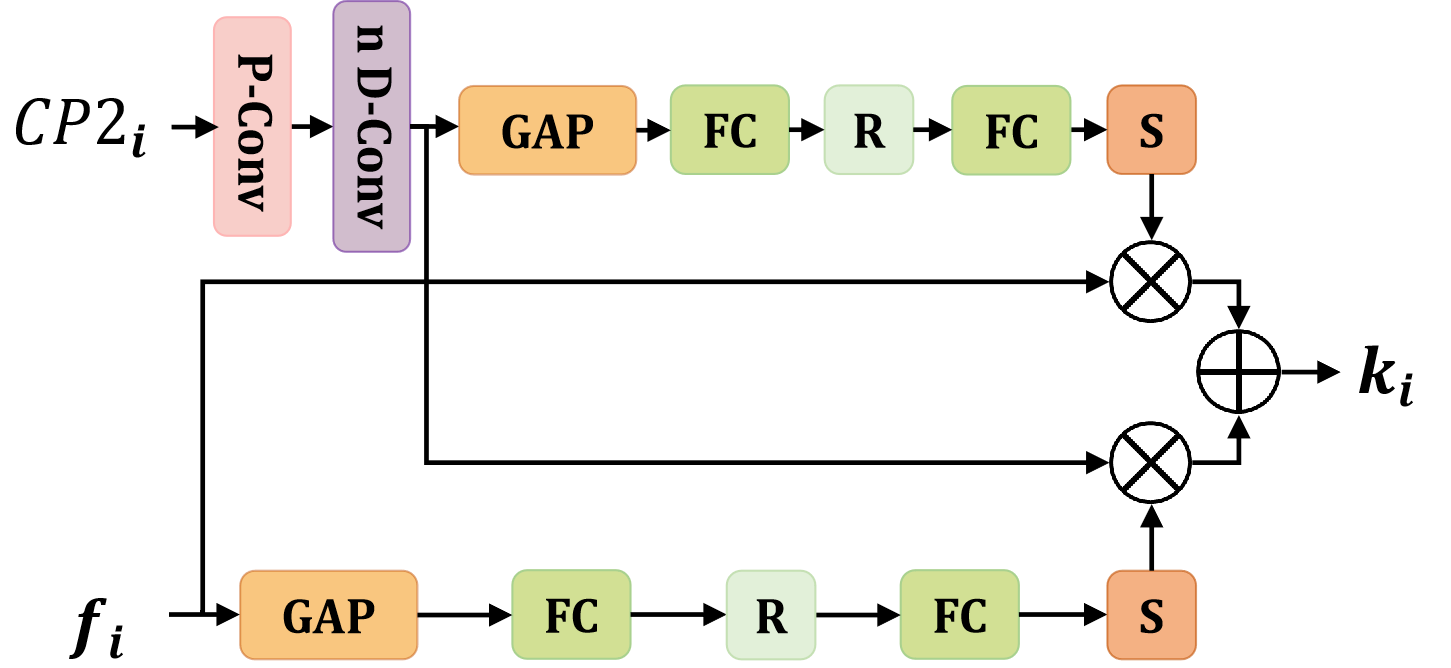}}
   \caption{ \small{Architectures for the ablation study on CHKIM. (a) Top-down modulation with point-wise channel
attention module (Local). (b) Bi-directional modulation with point-wise channel attention module (BiLocal).
(c) Bi-directional modulation with global channel attention
module (BiGlobal).}\label{abs_feature_fusion}}
\end{figure*}

\textit{2) Effect of DBPEA:} 
DBPEA is designed to collaboratively embed the CP components into the deep learning network, thereby enhancing the model's sensitivity to the location of small targets and its precise delineation of target contours. To investigate the benefits brought by this module, we compare the detection performance when using the entire architecture, using only the first branch to embed CP1, and using only the second branch to embed CP2. Table \ref{Ablation_DBPEA_1} compares their detection performance, including $IoU\left ( \% \right ) $, $F_{1}\left ( \% \right ) $, $P_{d}\left ( \% \right ) $, and $F_{a}\left ( \times 10^{-6}  \right ) $.





It can be observed that when CP1 or CP2 is introduced individually, the performance of the network is enhanced compared to the backbone, indicating that embedding prior information into deep learning models is effective. On the other hand, when both structures are introduced, that is, when the complete DBPEA module is utilized, the network performs even better, especially achieving the best values in  $IoU$, $F_{1}$, $P_{d} $, across all three datasets. This demonstrates that our designed DBPEA module can effectively assist the backbone network in extracting the contour details of small targets.

Additionally, to demonstrate that CP1 and CP2 can effectively extract contour information of small targets, we use images from two scenarios. As shown in Fig. \ref{PK_feature_1} and Fig. \ref{PK_feature_2}, the first column displays the original images, $I_{in}$ represents the heatmap of the original image, $\mathcal{F} _{c} \left ( I_{in},CP1 \right ) $ is the heatmap after fusing $I_{in}$ with $CP1$, $e_{0}$ to $e_{2}$ denote the heatmaps at each layer of the encoding process in DNIM, $f_{0}$ to $f_{3}$  represent the heatmaps at each layer of the decoding process, $k_{0}$ to $k_{3}$  indicate the heatmaps output at each layer after fusing $f_{i}$ with $CP2_{i}$ through CHKIM, $F$ represents the heatmap output by AGFEM, and the last two columns are the output detection results and GT, respectively. The color intensity varies from weak to strong, indicating energy levels from low to high. It is evident that when facing multiple targets, fusing position information $CP1$ with $I_{in}$ before feature extraction can detect dim targets against similar gray backgrounds, reducing the occurrence of missed detections and also suppress background clutter to some extent, as seen in the first and second rows in Fig. \ref{PK_feature_1}. Moreover, fusing $f_{i}$ with $CP2_{i}$ can significantly reduce clutter information outside the target area, helping the network focus on target-like regions and reduce false alarms, as illustrated in the first row and third row, columns 6 to 9 in Fig. \ref{PK_feature_1}. The introduction of $CP1$ and $CP2_{i}$ also aids in extracting detailed information from the images, better recovering the shapes of small targets, such as image in Fig. \ref{PK_feature_2}. The concentrated energy on the CSPENet heatmap and the detection results confirm the effectiveness of DBPEA design.

\textit{3) Effect of CHKIM:} CHKIM is utilized to infuse low-level structural information into deep features to achieve feature fusion, thereby enriching the information representation and enhancing the robustness of the model. To investigate the effectiveness of this module, we compare the performance of  CHKIM with three variants illustrated in Fig. \ref{abs_feature_fusion}, and the comparison results are shown in Table \ref{Ablation_CHKIM}.

Firstly, we compare the one-directional top-down modulation module, i.e., Local (in Fig. \ref{subfig_abs_fusion1}), with the two-directional modulation modules, i.e., BiLocal (in Fig. \ref{subfig_abs_fusion2}) and BiGlobal (in Fig. \ref{subfig_abs_fusion3}). It can be observed from Table \ref{Ablation_CHKIM} that both two-directional modulation modules consistently outperformed the one-directional modulation module across all metrics. Subsequently, we compare the symmetric feature fusion modules, BiLocal and BiGlobal, with the proposed CHKIM to verify the effectiveness of asymmetric attention modulation. As shown in Table \ref{Ablation_CHKIM}, compared to the modulation schemes where channel attention scales are either local (BiLocal) or global (BiGlobal), the proposed CHKIM performed the best, improving $IoU$, $F_{1}$, and $P_{d}$ by at least $2.73\%$, $0.97\%$, and $0.21\%$ respectively, and reducing $F_{a}$ by at least $2.88\times 10^{-6}$ compared with other methods on NUDT-SIRST. Similar results are also achieved on IRSTD-1K and NUAA-SIRST. This demonstrates that CHKIM can exchange multi-layer information and produce richer feature representations.

\begin{table*}[htbp!]
    \centering
    \scriptsize
    \caption{\\\footnotesize \scshape The $IoU\left ( \% \right ) $, $F_{1}\left ( \% \right ) $, $P_{d}\left ( \% \right ) $, AND $F_{a}\left ( \times 10^{-6}  \right ) $ \scshape Values \scshape Achieved \scshape By \scshape The \scshape Main \scshape Variants \scshape Of \scshape Chkim \scshape  \scshape On \scshape Nudt-Sirst, \scshape Irstd-1k, \scshape And \scshape Nuaa-Sirst.\scshape The \scshape Best \scshape Results \scshape Are \scshape In \scshape Red, \scshape And \scshape The \scshape Second \scshape Best \scshape Results \scshape Are \scshape In \scshape Blue.
}
    \label{Ablation_CHKIM}
     \begin{tabular*}{\textwidth}{@{\extracolsep\fill}l|cccc|cccc|cccc}
    \hline
     \multirow{2}{*}{ Methods}  & \multicolumn{4}{c|}{ NUDT-SIRST \cite{li2022dense}}& \multicolumn{4}{c|}{IRSTD-1K \cite{zhang2022isnet} }& \multicolumn{4}{c}{NUAA-SIRST \cite{dai2021asymmetric} } \\\cline{2-13}
       & $IoU$↑  & $F_{1}$↑& $P_{d}$↑& $F_{a}$↑& $IoU$↑ & $F_{1}$↑& $P_{d}$↑& $F_{a}$↓& $IoU$↑ & $F_{1}$↑& $P_{d}$↑& $F_{a}$↑ \\\hline
      Local &87.18	&92.95	&97.46	&17.58	&62.87	&76.82	&90.57	&32.61	&73.90	&85.32	&93.91	&28.46 \\
       BiLocal &90.39	&{\color{blue}96.03}	&98.20	&11.79	&{\color{blue}65.23}	&{\color{blue}79.69}	&{\color{blue}92.58}	&26.61	&75.27	&{\color{blue}87.01}	&96.20	&19.36 \\
     BiGlobal  &{\color{blue}91.45}	&95.92	&{\color{blue}98.31}	&{\color{blue}7.80}	&64.17	&79.06	&92.25	&{\color{red}23.16}	&{\color{blue}76.58}	&86.93	&{\color{blue}96.58}	&{\color{blue}18.44}\\\hline
      \textbf{CHKIM}& {\color{red}94.18}  & {\color{red}97.00}  & {\color{red}98.52} &{\color{red}4.92}  &{\color{red}66.79}  &{\color{red}81.10} &{\color{red}93.27}  &{\color{blue}23.53}  &{\color{red}79.83} &{\color{red}87.41} &{\color{red}96.96} &{\color{red}15.18} \\\hline
    \end{tabular*}
\end{table*}

\begin{table*}[htbp!]
    \centering
    \scriptsize
    \caption{\\\footnotesize \scshape The $IoU\left ( \% \right ) $, $F_{1}\left ( \% \right ) $, $P_{d}\left ( \% \right ) $, AND $F_{a}\left ( \times 10^{-6}  \right ) $ \scshape Values \scshape Achieved \scshape By \scshape The \scshape Main \scshape Variants \scshape Of \scshape Agfem \scshape  \scshape On \scshape Nudt-Sirst, \scshape Irstd-1k, \scshape And \scshape Nuaa-Sirst. \scshape The \scshape Best \scshape Results \scshape Are \scshape In \scshape Red, \scshape And \scshape The \scshape Second \scshape Best \scshape Results \scshape Are \scshape In \scshape Blue.
}
    \label{Ablation_AGFEM}
     \begin{tabular*}{\textwidth}{@{\extracolsep\fill}ccc|cccc|cccc|cccc}
    \hline
     \multirow{2}{*}{ CA} &\multirow{2}{*}{ SA} &\multirow{2}{*}{ Residual}  & \multicolumn{4}{c|}{ NUDT-SIRST \cite{li2022dense}}& \multicolumn{4}{c|}{IRSTD-1K \cite{zhang2022isnet} }& \multicolumn{4}{c}{NUAA-SIRST \cite{dai2021asymmetric} } \\\cline{4-15}
       & & & $IoU$↑  & $F_{1}$↑& $P_{d}$↑& $F_{a}$↑& $IoU$↑ & $F_{1}$↑& $P_{d}$↑& $F_{a}$↑& $IoU$↑ & $F_{1}$↑& $P_{d}$↑& $F_{a}$↑ \\\hline
      \textbf{ $\times $ } & \textbf{ $\times $ } &\textbf{ $\times $ } &93.28	&95.13	&96.62	&7.24	&65.23	&79.77	&91.47	&34.63	&78.51	&86.51	&95.82	&24.97 \\
      \textbf{ $\surd $ } &\textbf{ $\surd $ } &\textbf{ $\times $ }  &93.41	&95.92	&{\color{blue}97.88}	&6.31	&66.02	&{\color{blue}80.68}	&91.57	&30.28	&78.96	&86.79	&96.20	&22.35\\
       \textbf{ $\surd $ } &\textbf{ $\times $ } &\textbf{ $\surd $ } &93.70	&96.23	&96.67	&6.03	&{\color{blue}66.29}	&80.59	&{\color{blue}92.93}	&29.45	&{\color{blue}79.21}	&{\color{blue}87.79}	&{\color{blue}96.58}	&{\color{blue}16.34} \\
       \textbf{ $\times $ } &\textbf{ $\surd $ }  &\textbf{ $\surd $ } &{\color{blue}94.15}	&{\color{blue}96.56}	&97.57	&{\color{blue}5.29}	&66.20	&80.44	&92.26	&{\color{red}22.58}	&79.02	&87.01	&{\color{red}96.96}	&16.91 \\     \hline
      $\boldsymbol{\surd}$ &$\boldsymbol{\surd}$  &$\boldsymbol{\surd}$ &{\color{red}94.18}  &{\color{red}97.00}  & {\color{red}98.52} &{\color{red}4.92}  &{\color{red}66.79}  &{\color{red}81.10} &{\color{red}93.27}  &{\color{blue}23.53}  &{\color{red}79.83} &{\color{red}87.41} &{\color{red}96.96} &{\color{red}15.18} \\\hline
    \end{tabular*}
\end{table*}

\textit{4) Effect of AGFEM:} AGFEM is used for adaptive feature enhancement, integrating shallow features rich in spatial information with deep features rich in semantic information to generate more robust feature maps as output. To investigate the benefits brought by this module, we compare the performance with and without AGFEM, as well as three variants of AGFEM, with the results shown in Table \ref{Ablation_AGFEM}.

Compared to the network without AGFEM, the addition of AGFEM led to improvements in almost all metrics across the three datasets. Specifically, $IoU$ increases by $0.9\%$, $1.56\%$, and $1.32\%$ on NUDT-SIRST, IRSTD-1K, and NUAA-SIRST, respectively, while $F_{1}$ increases by $0.04\%$, $2.24\%$, and $1.48\%$, and $P_{d}$ improves by $1.9\%$, $1.8\%$, and $1.14\%$. Concurrently, $F_{a}$ also improves. In addition, if the residual structure or spatial attention or channel attention is removed, $IoU$, $F_{1}$, $P_{d}$ all decrease while $F_{a}$ increases compared to using the proposed AGFEM. These results confirm that AGFEM contributes positively to the overall performance.

\section{Conclusion}
This paper proposes CSPENet, a  novel infrared target detection network. By embedding two collaborative priors into the network, CSPENet effectively mitigates performance degradation caused by the limitations of existing methods in terms of target localization accuracy and contour detail description. Specifically, we design SCPEM to generate two collaborative prior components: one encoding  boosted saliency  spatial localization constraints, and the other capturing  multi-level contour detail features of the target. Building on this, DBPEA establishes differentiated feature fusion pathways, embedding these  two priors at optimal positions of the network to enhance target localization and structural detail fidelity. Additionally, we develop AGFEM to improve the feature utilization by integrating multi-scale salient features. Extensive experiments on public datasets demonstrate that our network significantly improves detection accuracy, especially in challenging scenarios involving  irregular target contour, complex background noise, and low signal-to-noise ratios. 

While CSPENet has shown promising results, our current validation remains limited to implementation on a popular  segmentation backbone network. Our future work will aim to verify the generalizability of our designs across diverse backbone architectures. Moreover, the growing adoption of infrared images in practical applications has revealed two critical challenges requiring attention: the detection of subpixel-scale tiny targets, and the identification of objects with unstructured contour characteristics. We intend to conduct comprehensive evaluations on these datasets once they become available. In addition, with the emergence of multi-frame datasets, the development of a spatio-temporal joint modeling framework that leverages multi-frame temporal information to enhance dynamic target detection performance would be another potential direction for exploration.


%




\bibliographystyle{IEEEtran}
\bibliography{reference} 

\ifCLASSOPTIONcaptionsoff
  \newpage
\fi

\begin{IEEEbiography}[{\includegraphics[width=1in,height=1.25in,clip,keepaspectratio]{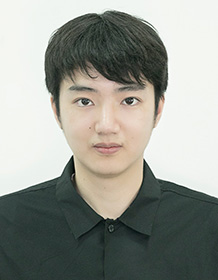}}]{Jiakun Deng} received the M.S. degree in Optical Engineering from the School of Optoelectronic Science and Engineering, University of Electronic Science and Technology of China (UESTC), Chengdu, China, in 2022. He is currently pursuing the Ph.D. degree in Electronic Information with the School of Information and Communication Engineering, UESTC. His research interests include computer vision, infrared small target detection and tracking, and object recognition.
\end{IEEEbiography}

\begin{IEEEbiography}[{\includegraphics[width=1in,height=1.25in,clip,keepaspectratio]{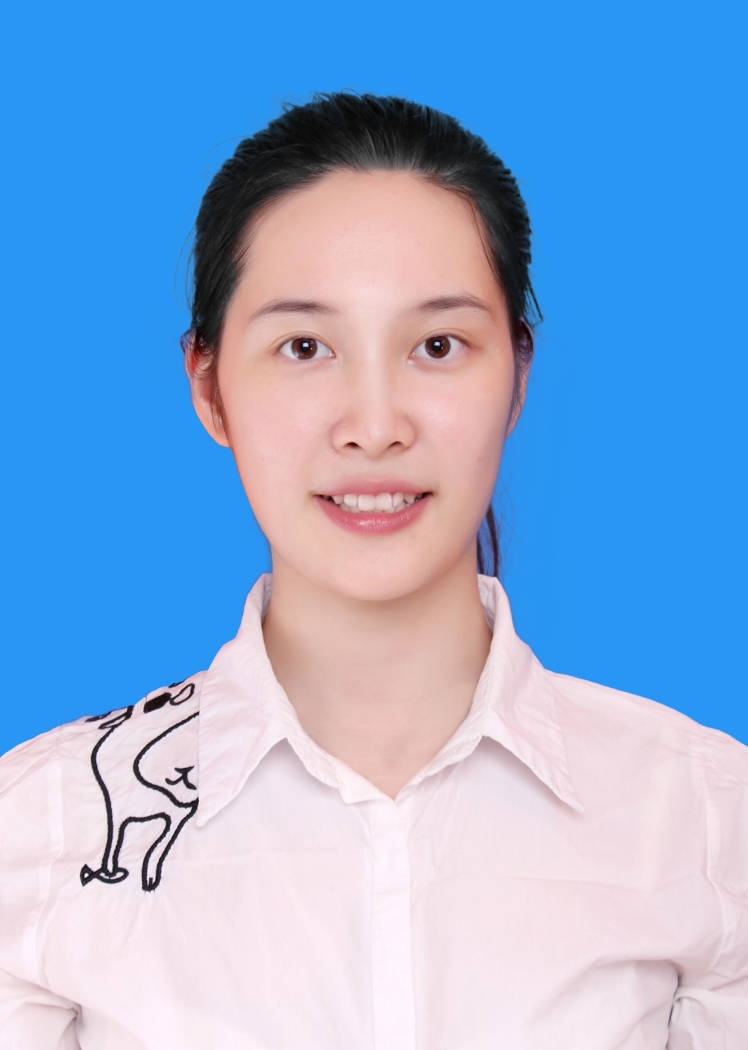}}]{KeXuan Li} received the B.S. degree from Huaqiao University, Quanzhou, China in 2023. She is currently working toward the M.E. degree in electronic information with the School of Information and Communication Engineering, University of Electronic Science and Technology of China, Chengdu, China. Her research interests include image processing, computer vision, and infrared small target detection.
\end{IEEEbiography}

\begin{IEEEbiography}[{\includegraphics[width=1in,height=1.25in,clip,keepaspectratio]{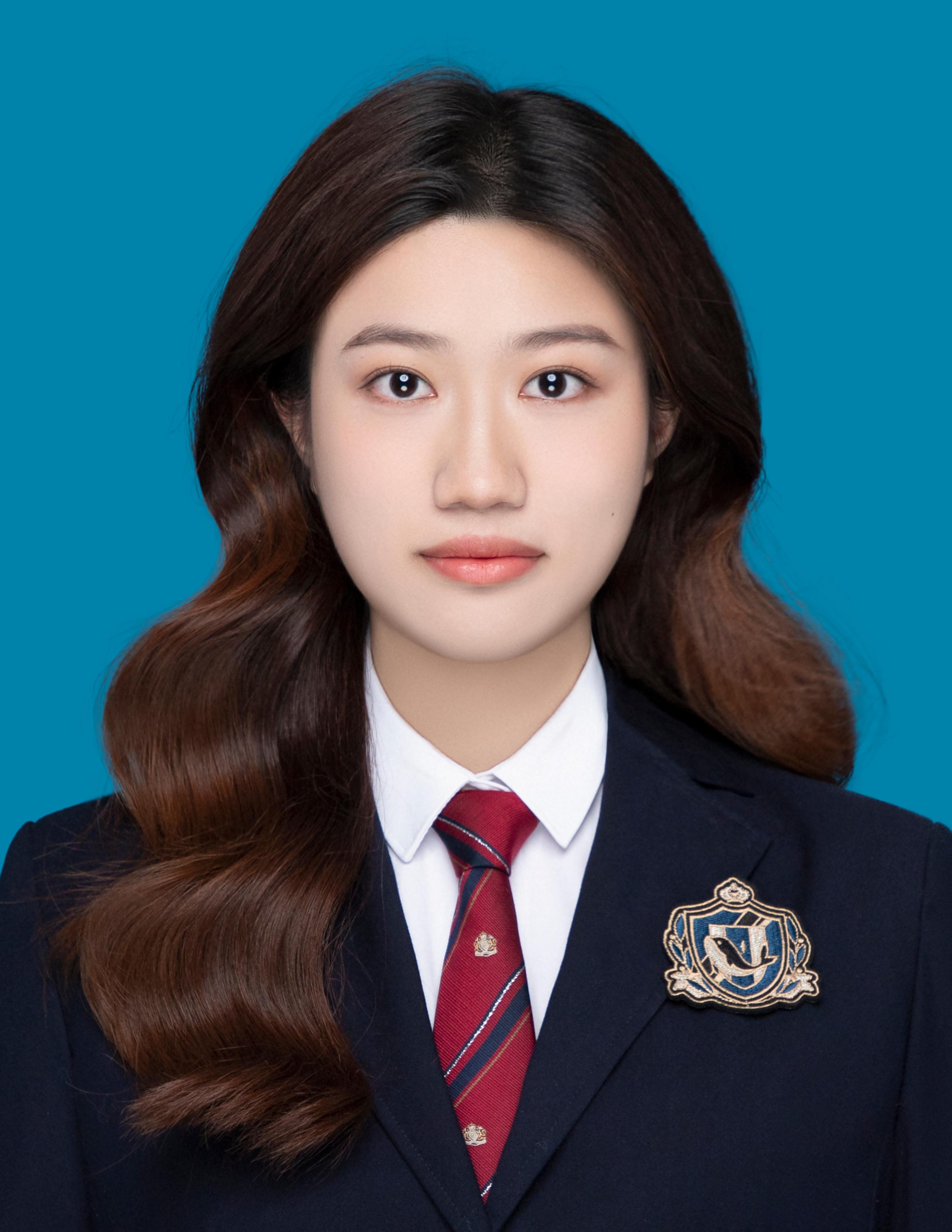}}]{Xingye Cui} received the B.E. degree in communication engineering from the School of Computer Science, Jiangsu University of Science and Technology, Zhenjiang, China, in 2023. She is currently working toward the M.E. degree in electronic information with the School of Information and Communication Engineering, University of Electronic Science and Technology of China, Chengdu, China. Her research interests include image processing, computer vision, and infrared small target detection.
\end{IEEEbiography}

\begin{IEEEbiography}[{\includegraphics[width=1in,height=1.25in,clip,keepaspectratio]{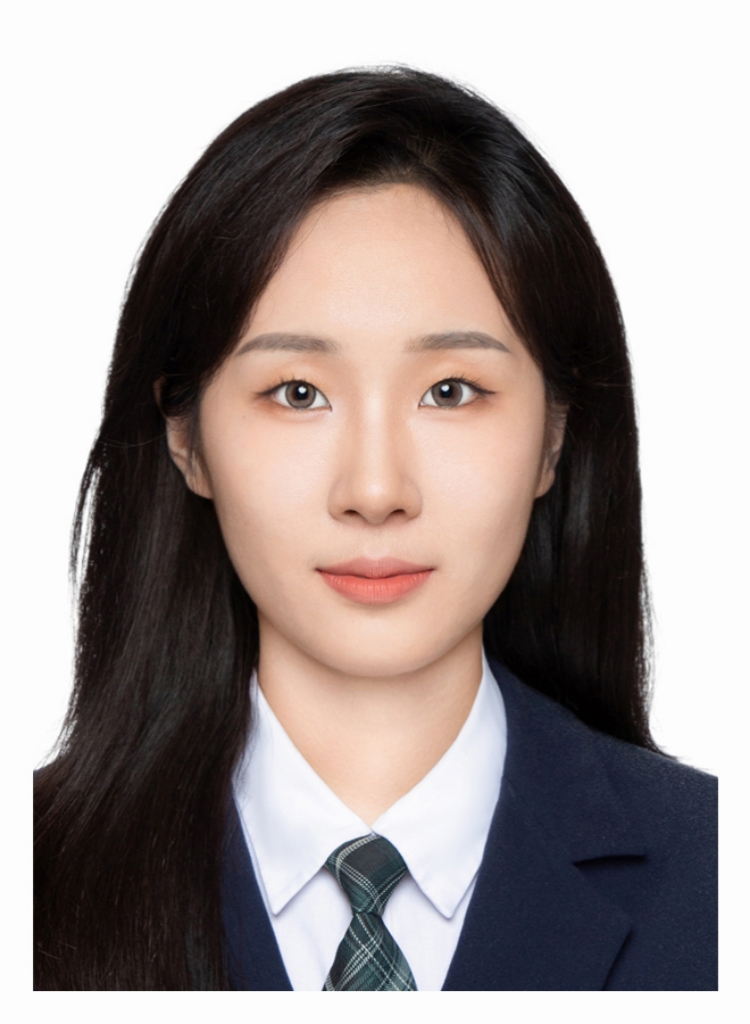}}]{Jiaxuan Li} graduated from Wuhan University of Technology in 2024 with a Bachelor's degree. She is currently pursuing a Master's degree at the University of Electronic Science and Technology of China in Chengdu, China. Her research interests primarily focus on the fields of target recognition, image processing, and computer vision.
\end{IEEEbiography}

\begin{IEEEbiography}[{\includegraphics[width=1in,height=1.25in,clip,keepaspectratio]{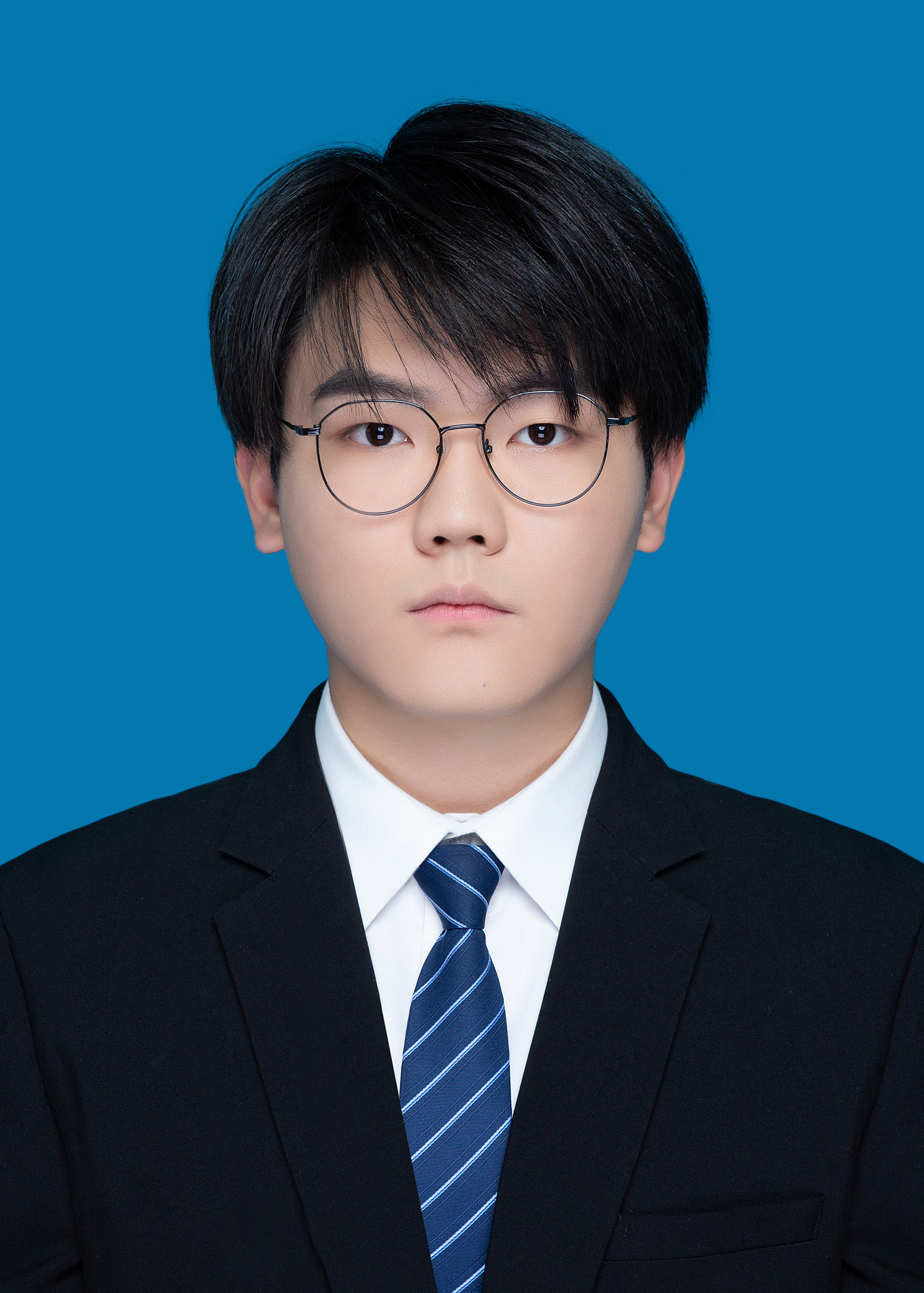}}]{Chang Long} received the B.E. degree from the School of Information and Communication Engineering, University of Electronic Science and Technology of China (UESTC), Chengdu, China, in 2023. He is currently working toward the M.E. degree in information and communication engineerng with the School of Information and Communication Engineering, UESTC, Chengdu, China.  His research interests include image processing, computer vision, and infrared small target detection.
\end{IEEEbiography}

\begin{IEEEbiography}[{\includegraphics[width=1in,height=1.25in,clip,keepaspectratio]{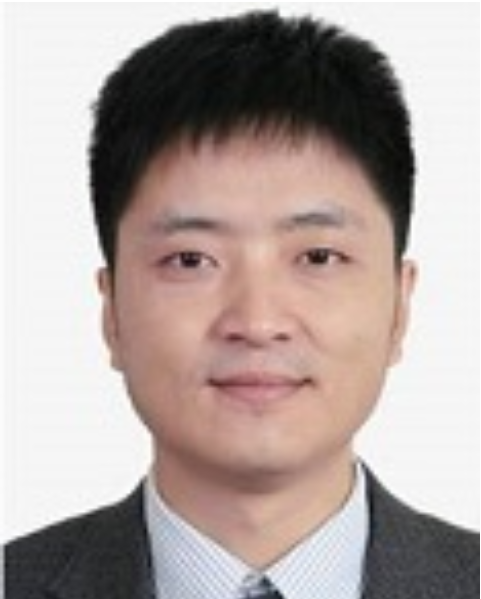}}]{Tian Pu} graduated from Wuhan University of Technology in 2024 with a Bachelor's degree. She is currently pursuing a Master's degree at the University of Electronic Science and Technology of China in Chengdu, China. Her research interests primarily focus on the fields of target recognition, image processing, and computer vision.
\end{IEEEbiography}

\begin{IEEEbiography}[{\includegraphics[width=1in,height=1.25in,clip,keepaspectratio]{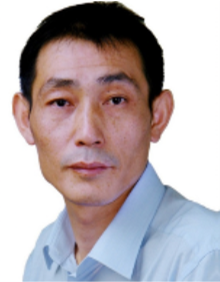}}]{Zhenming Peng} (Member, IEEE) received his Ph.D. degree in geodetection and information technology from the Chengdu University of Technology, Chengdu, China, in 2001. From 2001 to 2003, he was a post-doctoral researcher with the Institute of Optics and Electronics, Chinese Academy of Sciences, Chengdu, China. He is currently a Professor with the University of Electronic Science and Technology of China, Chengdu. His research interests include image processing, machine learning, objects detection and remote sensing applications. Prof. Peng is members of many academic organizations, such as Institute of Electrical and Electronics Engineers (IEEE), Optical Society of America (OSA), China Optical Engineering Society (COES), Chinese Association of Automation (CAA), Chinese Society of Astronautics (CSA), Chinese Institute of Electronics (CIE), and China Society of Image and Graphics (CSIG), etc.
\end{IEEEbiography}




\end{document}